\documentclass[10pt,twocolumn,letterpaper]{article}

\usepackage{wacv}
\usepackage{times}
\usepackage{epsfig}
\usepackage{graphicx}
\usepackage{amsmath}
\usepackage{amssymb}

\usepackage{multirow}
\usepackage{algorithm}
\usepackage{algpseudocode}
\usepackage{makecell}
\usepackage{enumitem}
\usepackage{pifont}
\usepackage{array}

\usepackage{color}
\usepackage{subfig}
\usepackage{comment}
\usepackage{float}

\usepackage{booktabs}
\usepackage{enumitem}
\usepackage{tabularx}
\usepackage{multirow}
\usepackage{stfloats}
\usepackage{threeparttable}
\usepackage[usenames,dvipsnames]{xcolor}

\newcommand{\cmark}{\ding{51}}%
\newcommand{\xmark}{{\color{red} \ding{55}}}%

\DeclareMathOperator*{\argmin}{arg\,min}
\DeclareMathOperator*{\rank}{Rank}

\newcolumntype{P}[1]{>{\centering\arraybackslash}p{#1}}

\newcommand{\stacmr}{StacMR}

\def\wacvPaperID{707} 

\wacvfinalcopy 
\newif\ifarxiv
\arxivtrue 
\ifwacvfinal
\def\assignedStartPage{1} 
\fi
\usepackage[pagebackref=true,breaklinks=true,bookmarks=false]{hyperref}

\ifwacvfinal
\else
\fi

\begin{document}

\title{\stacmr : Scene-Text Aware Cross-Modal Retrieval}
\author{Andr\'{e}s Mafla$^\dag$
\quad
Rafael S. Rezende$^\ddag$
\quad
Llu\'{i}s G\'{o}mez$^\dag$
\quad
Diane Larlus$^\ddag$
\quad
Dimosthenis Karatzas$^\dag$\\
$\dag$~Computer Vision Center, Universitat Aut\'{o}noma de Barcelona \quad $\ddag$~NAVER LABS Europe
}

\maketitle

\begin{abstract}

Recent models for cross-modal retrieval have benefited from an increasingly rich understanding of visual scenes, afforded by scene graphs and object interactions to mention a few. This has resulted in an improved matching between the visual representation of an image and the textual representation of its caption. Yet, current visual representations overlook a key aspect: the text appearing in images, which may contain crucial information for  retrieval. In this paper, we first propose a new dataset that allows exploration of cross-modal retrieval where images contain scene-text instances. Then, armed with this dataset, we describe several approaches which leverage scene text, including a better scene-text aware cross-modal retrieval method which uses specialized representations for text from the captions and text from the visual scene, and reconcile them in a common embedding space. Extensive experiments confirm that cross-modal retrieval approaches benefit from scene text and highlight interesting research questions worth exploring further. Dataset and code are available at
\url{http://europe.naverlabs.com/stacmr}.
\end{abstract}

\vspace{-3mm}
\section{Introduction}

\begin{figure}[t!]
    \centering
    \includegraphics[width=\linewidth]{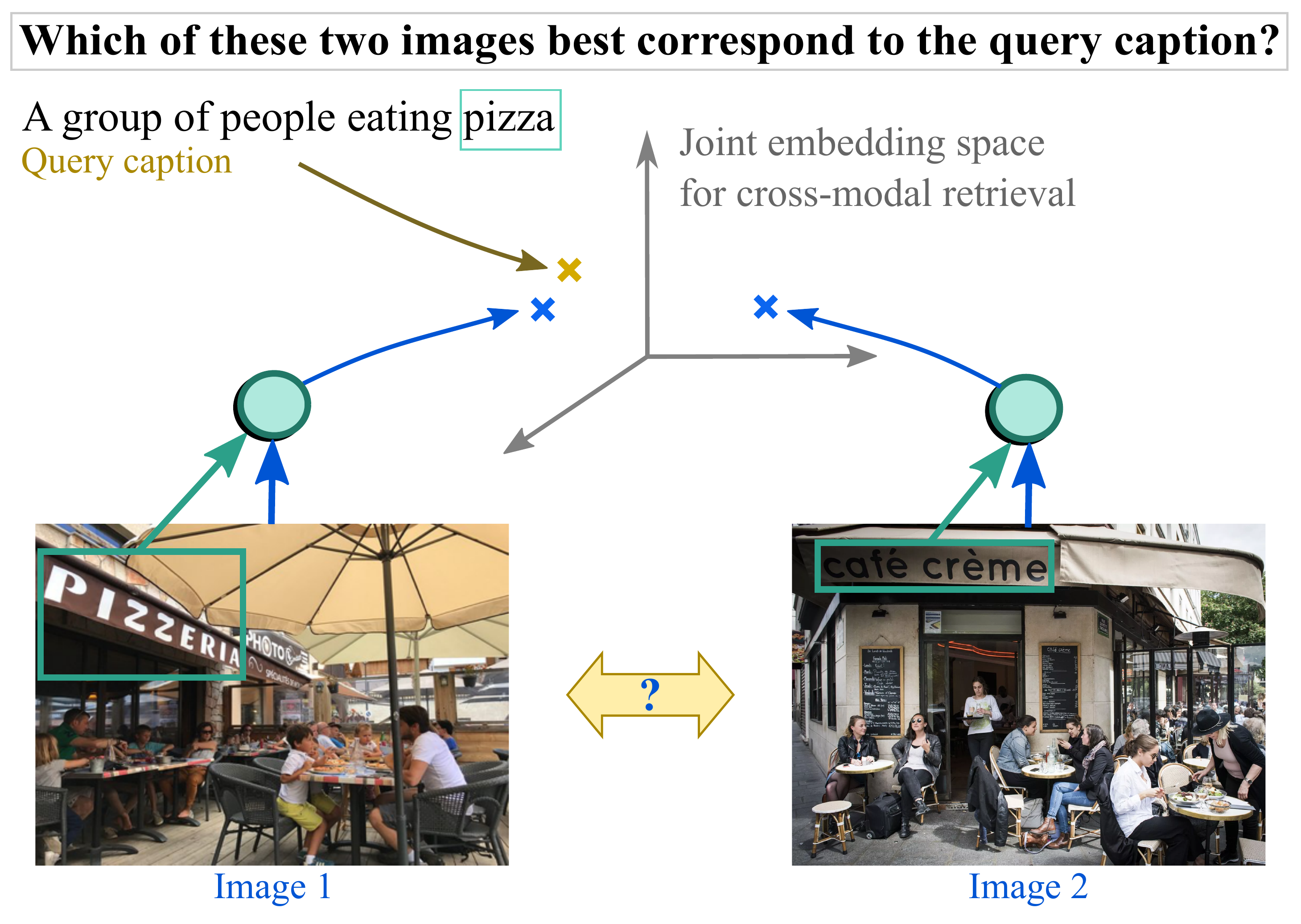}
    \caption{This paper introduces the \textbf{scene-text aware cross-modal retrieval}
       (\stacmr) task and 
      studies scene text as a third modality for cross-modal retrieval.
      For the example query above, the restaurant name provides crucial information to disambiguate two otherwise equally relevant results.} \label{fig:stacmr}
\end{figure}

Textual content is omnipresent in most man-made environments and plays a 
crucial role as it conveys key information to understand a visual scene.
Scene text commonly appears in natural images, especially in urban
scenarios, for which about half of the images habitually contain text~\cite{veit2016coco}. This is especially relevant when considering vision
and language tasks, and in particular, related to our work, cross-modal retrieval.
Scene text is 
a rich, explicit and semantic source of information
which can be used to disambiguate the fine-grained semantics of a visual scene
and can help to provide a suitable
ranking for otherwise equally probable results (see example in Figure~\ref{fig:stacmr}).
Thus explicitly taking advantage of this third modality should be a natural step
towards more efficient retrieval models. Nonetheless, and to the best of our
knowledge, scene text has never been used for the task of cross-modal retrieval,
and the community lacks a benchmark to properly address this research
question. Our work tackles these two open directions.

Scene text has been successfully leveraged to improve several semantics tasks in
the past, such as fine-grained image
classification~\cite{bai2018integrating,karaoglu2017text, mafla2020fine,movshovitz2015ontological}, visual question answering (VQA)~\cite{biten2019stvqa,singh2019towards} or image captioning~\cite{sidorov2020textcaps}.
Current mainstream methods tackle cross-modal retrieval by either learning to
project images and their captions into a joint embedding
space~\cite{faghri18vse,kiros2014unifying,  li2019visual, wang2020consensus}
or by directly comparing image regions and caption fragments to compute a
similarity score~\cite{karpathy2015deep, lee2018stacked}. Although significant
gaps have been overcome by previous methods, the lack of integration between scene
text and the other modalities still hinder a fuller image comprehension.
The intuition that serves as the foundation of this work stems from the notion
that scene text, found in natural images, can be exploited to obtain stronger
semantic relations between images and their captions. Obtaining such relations opens up the path toward improved retrieval systems in which scene text can serve as a guiding signal to provide more relevant and precise results.

This paper introduces the Scene-Text Aware Cross-Modal Retrieval (\stacmr)
task which aims to capture the interplay between captions, scene text, and visual signals.
To overcome the data scarcity of the proposed task, we have constructed a dataset based on COCO images~\cite{lin2014microsoft} which we name 
COCO-Text Captioned (CTC). 
It exhibits unique characteristics compared to other datasets employed for multi-modal tasks and does not share their bias towards scene text as the main component present in an image. 
In this work, we also evaluate the performance of different state-of-the-art cross-modal retrieval models, their limitations, and we propose distinctive baselines to solve this task. 

Concretely, the contribution of this paper is threefold.
First, we introduce a new task called Scene-Text Aware Cross-Modal Retrieval (or
\stacmr{} in short), as an extension to
cross-modal retrieval. In this task, leveraging the additional modality provided
by scene text is crucial to further reduce the heterogeneity gap between images and captions.

Second, we describe a new dataset, COCO-Text Captioned (CTC), as
the first dataset properly equipped to evaluate the \stacmr{} task. We highlight the
importance of the role that incidental scene text plays when interpreting an
image and its positive impact on retrieval results. We also compare the properties of our CTC dataset with similar existing datasets containing scene text and captions.

Finally, we provide a extensive analysis of
CTC. In particular (1) we benchmark the combination of different cross-modal baselines to model the interaction between scene text, visual, and caption information, and (2) we propose and evaluate a new model, STARNet, which explicitly learns to combine visual and scene-text cues into a unified image-level representation.

\section{Related Work}
\label{sec:rw}

\textbf{Scene-Text Detection and Recognition.}
Due to the large variance in text instances found in the
wild~\cite{chen2020text,zhu2016scene}, scene text detection and recognition is
still an active research field.
Methods such as EAST~\cite{zhou2017east}, Textboxes++~\cite{liao2018textboxes++} or LOMO~\cite{zhang2019look} draw inspiration from general object
detectors~\cite{he2017mask, liu2016ssd, redmon2016yolo9000, ren2015faster} and typically localize text instances by regressing pre-defined anchor boxes or pixels.

Moreover, pipelines trained end-to-end often benefit from both tasks, detection and recognition. Mask Textspotter~\cite{lyu2018mask} is an end-to-end segmentation-based approach which detects and recognizes text in arbitrary shapes. Similarly, \cite{he2018end} extracts image features with a CNN that are later refined by two Long-Short Term Memories (LSTMs) along with a text-alignment layer to perform these two tasks jointly.
In another approach, \cite{yu2020towards} employs a semantic reasoning network to mitigate transcriptions by projecting textual regions in a learned semantic space.

\textbf{Scene Text in Vision and Language.}
Methods for vision and language tasks typically align both
modalities and often perform visual reasoning.
Only recently have they started including scene text as an additional modality.
Works such as Text-VQA~\cite{singh2019towards} and Scene-Text
VQA~\cite{biten2019stvqa} focus on models capable of reading text in the wild
as well as reasoning about the inherent relations with visual features to
properly answer a question given in natural language. Scene text has also been
used to perform fine-grained image classification: 
\cite{bai2018integrating, karaoglu2017text, mafla2020multi} learn a shared semantic space between visual
features and text to perform classification while~\cite{mafla2020fine} uses the Pyramidal Histogram Of Characters (PHOC)~\cite{almazan2014word, Gomez_2018_ECCV, mafla2020real} descriptor as a way of overcoming OCR limitations and learn a morphological space suitable for the task.
Other works~\cite{gomez2018single,mishra2013image} perform scene-text based image search, where we query with a word and retrieve images containing such word.
Closer to our work, the TextCaps dataset~\cite{sidorov2020textcaps} includes scene text into textual descriptions. 
We discuss further the link with our work in Section~\ref{sec:dataset}.

\textbf{Cross-Modal Retrieval.}
Most cross-modal retrieval (CMR) approaches learn a joint representation
space together with visual and textual embedding functions which produce similar
representations for semantically related input, \eg an image and its captions. 
Often, the visual embedding function is a CNN and the textual one a
recurrent neural network~\cite{faghri18vse,ma2015multimodal,mao2014deep, wang2016learning}.
Other approaches use regions of interest given by a
detector~\cite{anderson2018bottom}. These approaches align each visual region
with a corresponding caption word to get a finer-grained image
representation~\cite{chen2020imram,karpathy2014deep, lee2018stacked, li2019visual,
  wang2020consensus,  zhang2020context}. Some methods also use attention mechanisms~\cite{lee2018stacked,nam2017dual,  song2019polysemous} that model detailed interactions between captions and image regions.
More recently, transformers~\cite{vaswani2017attention} have been
combined~\cite{tan2019lxmert,wei2020multi, wu2019learning} to perform
multi-layered self-attention operations in order to better align visual and textual features. Other works~\cite{li2019visual, wang2020cross} perform visual reasoning by employing graph convolutional networks~\cite{kipf2016semi} which yield a rich set of features by defining a relational graph between paired images and sentences. Closer to our work, Vo~\etal~\cite{vo2019composing} propose to use text modifiers along with images to retrieve relevant images. 

\section{The CTC Dataset}
\vspace{-1mm}
\label{sec:dataset}
This section introduces the proposed COCO-Text Captioned (CTC) dataset.
We first describe how it was gathered and tailored for the new \stacmr{} task, which extends traditional cross-modal retrieval to leverage information from a third modality: \textit{scene text}.
(Section~\ref{sec:dataset_collection}).
Then we present CTC statistics and discuss the dataset in the light of other benchmarks and in particular the most related dataset:
TextCaps~\cite{sidorov2020textcaps} (Section~\ref{sec:textcaps_comparison}).

\subsection{Data Collection and Statistics}
\label{sec:dataset_collection}

\textbf{Building the Dataset.}
A suitable dataset for the proposed \stacmr{} task requires the availability of
these three modalities: \textit{images, captions} and \textit{scene text}. The most commonly used
datasets for the cross-modal retrieval task~\cite{engilberge2018finding,faghri18vse, 
   klein2015associating, lee2018stacked,li2019visual, tan2019lxmert, wang2020consensus,
  wang2020cross} are COCO Captions~\cite{chen2015microsoft}, commonly known as MS-COCO in the cross-modal literature, and
Flickr30K~\cite{young2014image}. Only very few images from Flickr30K contain
scene text (see Table~\ref{table:text_instances}), so we decided to start from COCO Captions, a subset of the COCO
dataset~\cite{lin2014microsoft}. Additionally, the 
reading systems community commonly uses the COCO-Text
dataset~\cite{veit2016coco}.
It contains a sample of $63,686$ COCO images with fully annotated scene-text instances.
Among the COCO-Text images, we selected the ones that contain machine printed,
legible text in English, 
leading to a total of $17,237$ images.
In order to gather only images with the three modalities, 
we finally
select the intersection between the filtered COCO-Text and COCO Captions.
This leads to a multimodal dataset of $10,683$ items, each item
consisting of an image with scene text and five captions,
referred to as
as \textit{COCO-Text Captioned (CTC)}.

\begin{figure}
    \centering
    \includegraphics[width=\linewidth]{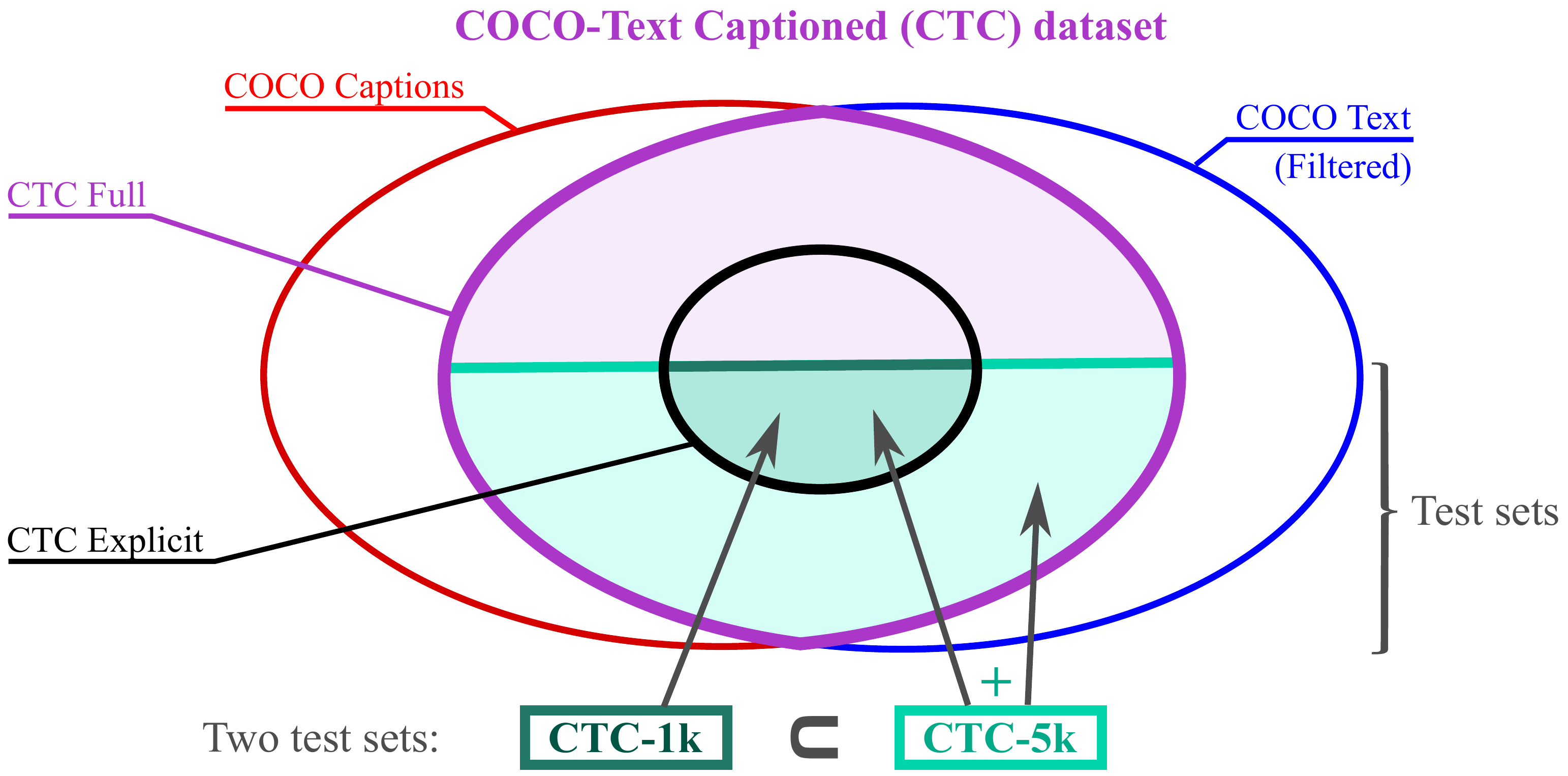}
    \caption{\textbf{Proposed CTC dataset}, which is 
      designed to allow a proper evaluation of the STACMR task, as all entries
      contain three modalities: image, scene text and caption. 
      }
    \label{fig:dataset}
\end{figure}

Note that the resulting CTC dataset shares $92.47\%$ of its images
with the original COCO caption training split. As a consequence, \textit{we can not use any models trained on COCO caption in our experiments}, as their training set would inevitably share images with our test set. The dataset's construction is
illustrated in Figure~\ref{fig:dataset}. 


\begin{table}[t!]
  \footnotesize
\begin{center}
\begin{tabular}{l|rrcc}
\toprule
\multirow{2}{*}{\textbf{Dataset}} & \multicolumn{1}{c}{\multirow{2}{*}{\textbf{\begin{tabular}[c]{@{}c@{}}Total \\ Images\end{tabular}}}} & \multicolumn{1}{c|}{\multirow{2}{*}{\textbf{\begin{tabular}[c]{@{}c@{}}Images \\ w/ Text\end{tabular}}}} & \multicolumn{2}{c}{\textbf{Annotations}} \\ \cline{4-5}
                                  & \multicolumn{1}{c}{}                                                                                  & \multicolumn{1}{c|}{}                                                                                    & Scene Text          & Captions           \\
                                  \hline
Flickr30K~\cite{young2014image}                & 31,783                                                                                                & 3,338$^\star$                                                                                                   & \xmark               & \cmark              \\
TextCaps~\cite{sidorov2020textcaps}                 & 28,408                                                                                                & 28,408$^\ddagger$                                                                                                  & \xmark               & \cmark              \\
COCO Captions~\cite{chen2015microsoft}                 & 123,287                                                                                               & 15,844$^\star$                                                                                                  & \xmark               & \cmark              \\
COCO-Text~\cite{veit2016coco}                & 63,686                                                                                                & 17,237$^\dagger$                                                                                                  & \cmark               & \xmark              \\
\midrule
\textbf{COCO-Text Caps}           & \textbf{10,683}                                                                                       & \textbf{10,683$^\dagger$}                                                                                         & \textbf{\cmark}      & \textbf{\cmark}    \\
\bottomrule
\end{tabular}
\end{center}
\caption{{\bf Datasets' statistics} for standard benchmarks and the proposed CTC. 
  $\dagger$ refers to COCO-Text filtered selecting machine printed, English and
  legible scene text only. $\star$ numbers obtained with method from ~\cite{mafla2020real}. $\ddagger$ numbers obtained with method from~\cite{borisyuk2018rosetta}.}
\label{table:text_instances}
\end{table}

\textbf{Statistics.}
Our only driver for building the CTC dataset has been to identify samples where all three modalities are available, without explicitly requiring at any point that scene text had any semantic relation to the captions. This is the most important requirement for a dataset where scene text is truly incidental and captions are not biased towards this additional modality. 
Despite this, to be coherent with the \stacmr{} task definition, it is paramount to show that the proposed CTC
dataset contains some inherent semantic relations between scene text found in an image and the captions that describe it.
To this end, we design three scenarios which illustrate this semantic relevance at
the \textit{image}, \textit{caption} and \textit{word}-level.


\begin{figure}[t!]
\begin{center}
\includegraphics[width=\linewidth, height=5.5cm]{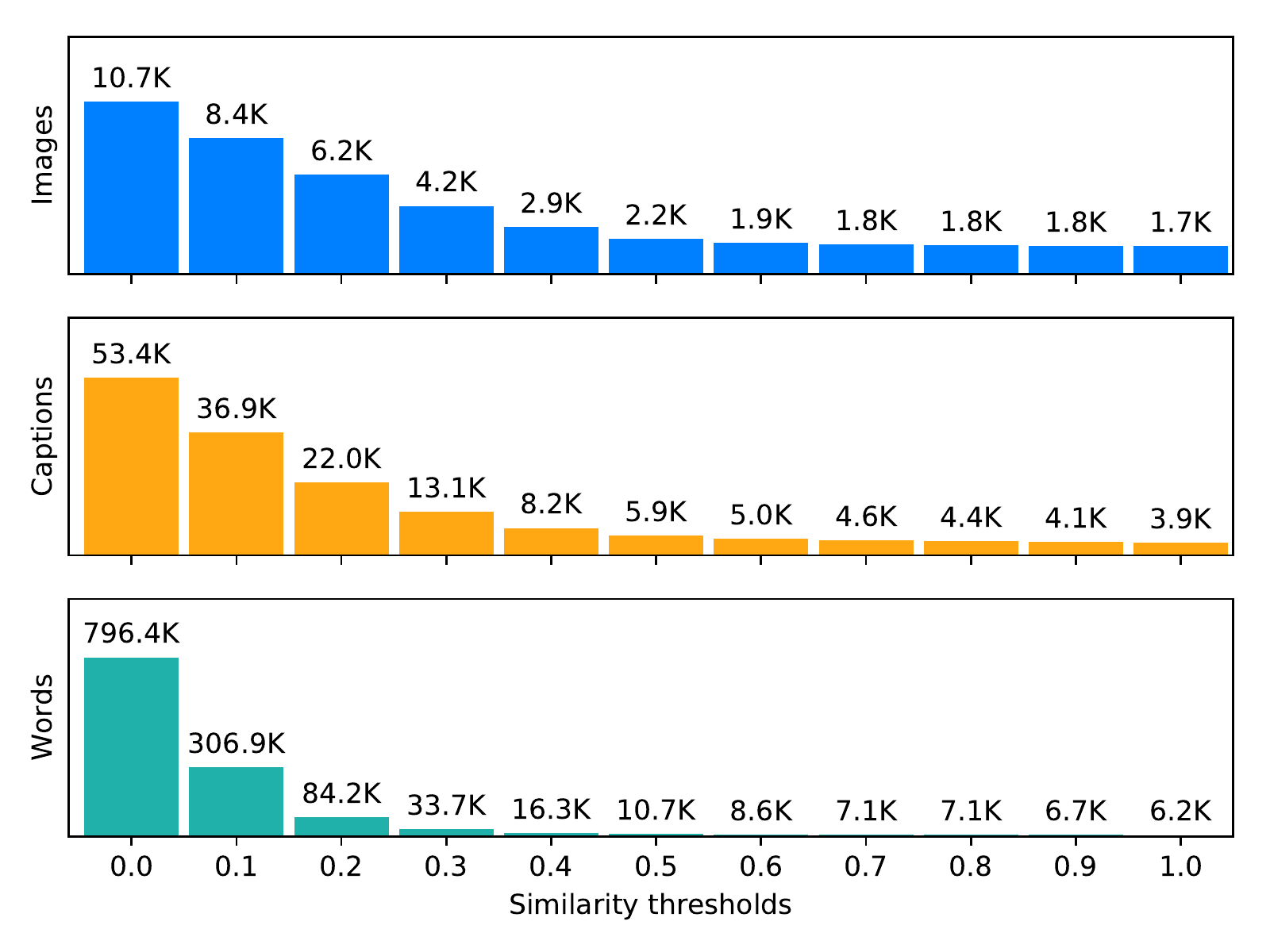}
\end{center}
\vspace{-0.8cm}
\caption{\label{fig:semantic_histogram} \textbf{CTC full statistics.} Cumulative histograms (as thresholds over similarity vary) of
    the semantic similarity between instances of scene-text tokens and a) all
    captions for an image (\textcolor{RoyalBlue}{Images}), b) individual captions (\textcolor{Dandelion}{Captions}), and c)
    individual words in captions (\textcolor{SeaGreen}{Words}).}

\end{figure}


More precisely, we first remove stop-words from captions and scene-text annotations, and
embed each remaining word with Word2Vec~\cite{mikolov2013distributed} vectors trained on the Google News dataset. The semantic
relevance between two words is defined as the cosine similarity between their
Word2Vec embeddings. 
We then consider three scenarios to showcase the relevance of scene text to image captions. 
The first scenario considers the highest semantic similarity between any scene-text word and any word from the set of $5$ captions, for each image. This scenario visualizes the \textit{semantic relation with images}, seen as sets of captions. 
The second scenario considers the highest semantic similarity between any scene-text word and any word from a corresponding caption. It highlights the \textit{semantic relation with individual captions}.
The third scenario considers how many
caption words are related to scene-text words.
This captures the \textit{semantic relation with individual words} in captions.

The three histograms of Figure~\ref{fig:semantic_histogram}
correspond to the three previously described scenarios. The fact that many words have a strong similarity at all three levels confirm that scene text can be used to model the semantics between the three studied modalities to further leverage them in order to obtain a better performing cross-modal retrieval system.

\begin{figure}[t]
  \setlength\extrarowheight{1pt} 
  \scriptsize
\begin{tabularx}{\linewidth}{@{\extracolsep{\fill}} cX}
  \toprule
  \textbf{Image} & \multicolumn{1}{c}{\textbf{Captions}} \\
\midrule
\multirow{5}{*}{\includegraphics[width=0.16\textwidth]{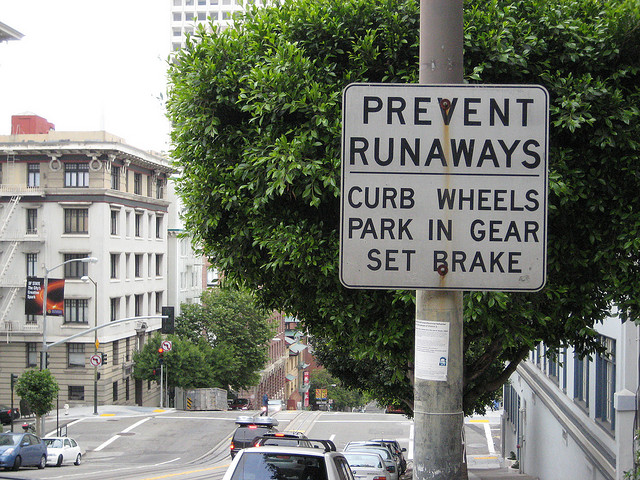}}
& {Sign warns against \textbf{runaway} vehicles along a hilly roadway.} \\
& {A white signing telling people how to \textbf{park} their cars on a steep hill.} \\
& {A sign explaining how to \textbf{park} on a hill is posted on the street.} \\
& {A warning sign is fastened to a post.} \\
& {Street sign with instructions on parking the hilly city roads.} \\
\midrule 
\multirow{5}{*}{\includegraphics[width=0.16\textwidth]{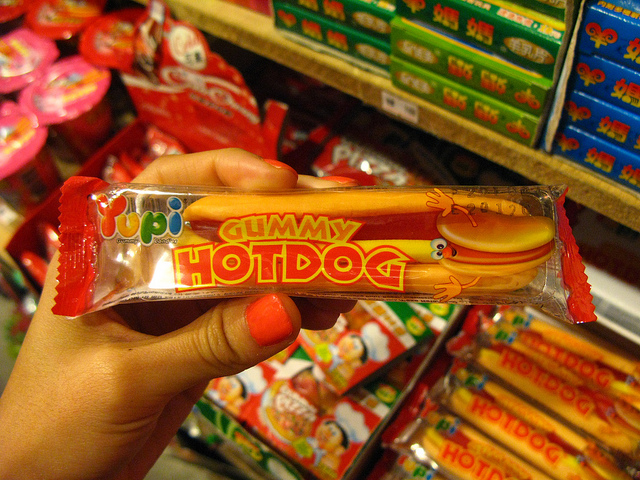}} 
 & {A person holding up a tasty looking treat.} \\
 & {A person holding up a \textbf{gummy hot dog} in their hand..} \\
 & {a closeup of a candy \textbf{gummy hot dog} in plastic packaging.} \\
 & {A \textbf{hotdog} that appears to be a \textbf{gummy hotdog.}} \\
 & {A \textbf{gummy hot dog} that is for sale.} \\
\midrule 
\multirow{5}{*}{\includegraphics[width=0.16\textwidth]{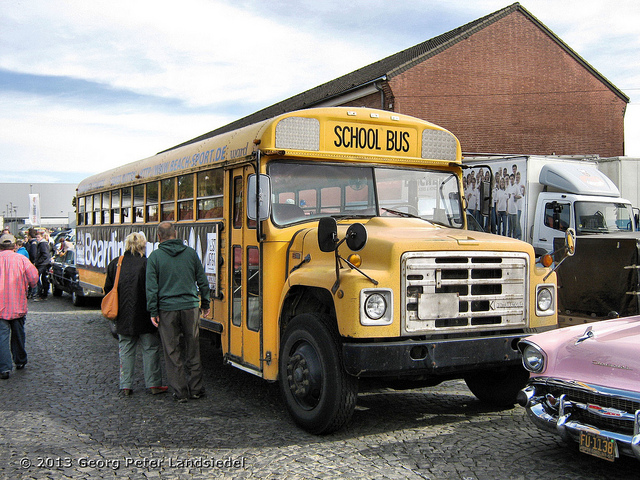}}  
  & {Parked \textbf{school bus} with a banner attached to it and people looking at it.} \\
  & {A man and a woman outside a \textbf{school bus}.} \\
  & {A \textbf{school bus} parked outside of a building.} \\
  & {A \textbf{school bus} sits parked as people walk by.} \\
  &{A \textbf{school bus} sitting on the side of the road near a pink car.} \\
\bottomrule
\end{tabularx}
\caption{\textbf{Image-caption pairs from the CTC dataset.} These
  images belong to CTC explicit, \ie their
  scene text and captions share at least one word  (marked in \textbf{bold}).
}
\label{fig:dataset_samples}
\end{figure}

As scene text provides fine-grained semantic information, its importance is query-dependant 
and it should be used selectively. 
An algorithm designed for the task should be able to
decide, for each image, to which extent scene text is relevant for the cross-modal retrieval task. 
In order to better capture this, we define two partitions of the CTC dataset. 
CTC presents a natural semantic split that is evident in Figure~\ref{fig:semantic_histogram} - a) that quantifies semantic similarity at the image-level.
The first quantization (threshold = $1$) corresponds to images for which at least
one word appears in both the scene text and one of the captions. We refer to this set
of $1,738$ images as \textit{CTC explicit}. We expect scene text from this set to often be relevant to the retrieval task. We employ the full CTC dataset, here referenced as \textit{CTC 
full} to avoid ambiguity, to evaluate 
the more generic scenario where the role of scene text for retrieval is a priori unknown. 
This second set contains the previously
mentioned explicit partition as well as images in which scene text is less
relevant according to the annotated captions.
Example image-caption pairs from \textit{CTC explicit} are shown in Figure~\ref{fig:dataset_samples}. This illustrates that
scene text provides a strong cue and fine-grained information for cross-modal retrieval.

For evaluation purposes, we define two test splits. 
The first one, which we refer to as
\textit{CTC-1K}, is a subset of \textit{CTC explicit}. The second test set, \textit{CTC-5K}, contains the previous $1,000$ explicit images of CTC-1K plus $4,000$ non-explicit images. The remaining $738$ explicit plus $4,945$ non-explicit images are used for training and validation purposes.

\subsection{Comparison with other Datasets}
\label{sec:textcaps_comparison}
Table~\ref{table:text_instances} provides a comparison with the previously mentioned datasets with statistics on the three modalities.  
Scene-text from COCO Captions~\cite{chen2015microsoft} and
Flickr30K~\cite{young2014image} was acquired using a scene-text
detector~\cite{mafla2020real}. As mentioned earlier, none of the existing
benchmarks contains samples where all three modalities are annotated.

\begin{figure}[t!]
\begin{center}
\includegraphics[width=\linewidth]{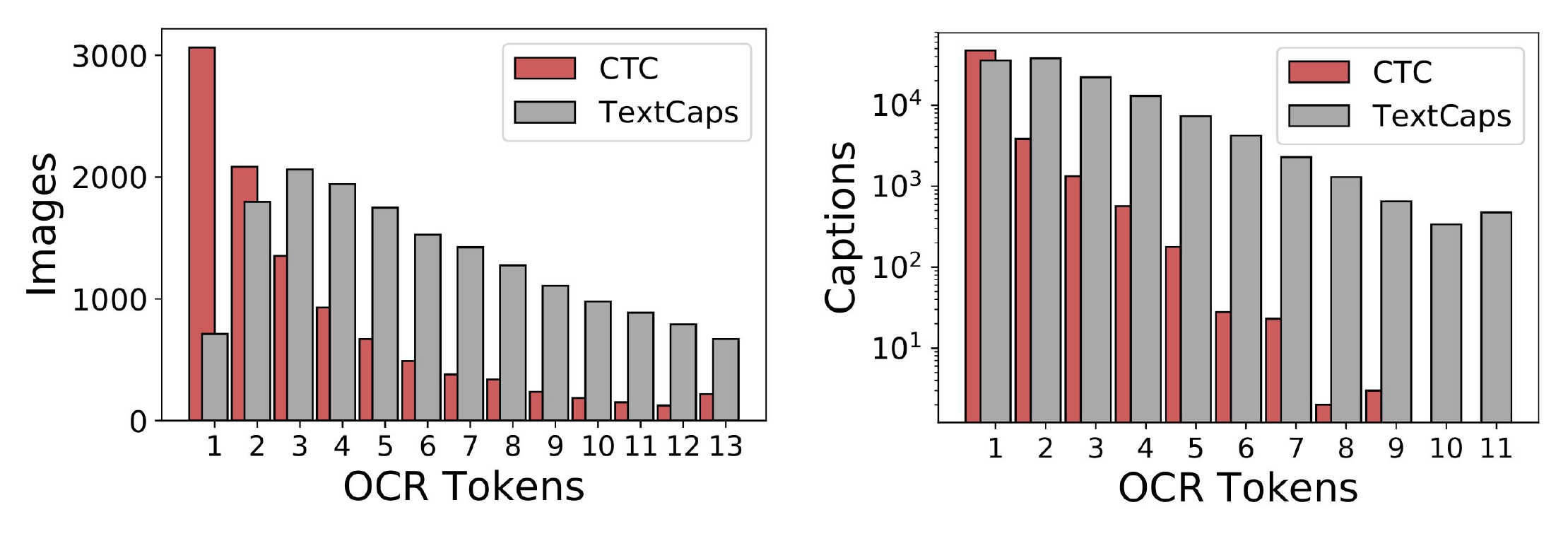}
\end{center}
\vspace{-3mm}
\caption{Histograms of the number of OCR tokens found in images (seen as sets of captions, left) and in individual captions (right) for the CTC and TextCaps datasets. 
}
\label{fig:comparison_histogram}
\end{figure}

Closely related to the proposed CTC dataset, TextCaps~\cite{sidorov2020textcaps} is an image captioning dataset that contains scene-text instances in all of its $28,408$ images. 
TextCaps is biased by design, as annotators were asked to describe an image in one sentence which would require reading the text in the image.
From the statistics shown in Figure~\ref{fig:comparison_histogram} it can be seen first, that TextCaps images were selected to contain more text tokens than should be naturally expected and second, that many more of these tokens end up being used in the captions compared to the unbiased captions of CTC.
The existing bias in TextCaps is also evident by analysing the intersection of $6,653$ images it has with the recently published Localized Narratives dataset~\cite{pont2019connecting}. 
From those $6,653$ images only $512$ ($10\%$) of them were annotated with captions that make use of any text tokens in the Localised Narratives dataset, where annotators were not instructed to always use the scene text.
According to our statistics, this is already higher than expected in the real world. This is because the Localised Narratives captions are long descriptions and tend to venture to fine-grained (localised) descriptions of images parts where text is more relevant.

The proposed CTC is a much less biased dataset in terms of caption generation. 
The objective is to provide realistic data that permit algorithms to explore the complex, real-life interaction between captions, visual and scene-text information, avoiding to assume or force any semantic relation between them.
More experiments showing the bias between TextCaps' captions and scene-text are provided in Section~\ref{sec:exp} and in the supplementary material.


\section{Method}
\label{sec:method}

This section describes approaches to tackle the \stacmr{} task. First, 
we propose strategies to directly apply standard pretrained cross-modal retrieval models to our new task and its three
modalities: images, captions and scene text (Section~\ref{sec:method_unsup}).
Second, we propose an architecture to learn a joint embedding space
for cross-modal retrieval in which the image embedding function learns to fuse both the visual
and the scene-text information (Section~\ref{sec:method_sup}).

\subsection{Re-Ranking Strategies}
\label{sec:method_unsup}

This subsection considers the image-to-caption retrieval task. Note that everything can easily be rewritten to consider the caption-to-image case. 

For \stacmr{}, images are multimodal objects: they contain visual information
as well as textual information coming from scene text. On the other hand, captions contain
textual information only. This asymmetry allows decomposing the \stacmr{} task
into two independent retrieval problems: visual-to-caption and
scene-text-to-caption. 
The first \textit{visual-to-caption} retrieval task performs comparisons between a purely visual
descriptor of the query image and the textual descriptor of the captions.
This corresponds to the standard cross-modal retrieval task as performed on Flickr30K or COCO Captions. 
The second, \textit{scene-text-to-caption} retrieval task, performs comparisons between the
textual descriptors of the scene text from the query image and the captions. Any textual descriptor could be used. In our experiments, we use the textual descriptor of a cross-modal retrieval
model as it has been tailored for capturing concepts relevant for images.

A pretrained cross-modal retrieval model relies on a metric space equipped with a similarity function which can indistinguishably compare visual and textual descriptors
and allows to rank all database elements according to a query.

\textbf{Notations.}
Given a query image $q$ and a caption from the gallery $d$, let 
$s_v(q,d)$ be the score between $q$ and $d$ using the image-to-caption
similarity from a cross-modal retrieval model and $s_t(q,d)$ the
 score between $q$ and $d$ using the scene-text-to-caption similarity from
that same model.

\textbf{Re-Ranking Strategies.}
The most straightforward way to obtain \stacmr{} results is simply to perform a
\textit{late fusion (LF)} of the ranking results obtained using both $s_v$ and $s_t$.
More formally, we compute the linear combination $s_{LT}$ of the scores $s_v$
and $s_t$, using a parameter $\alpha$:
\begin{align}
    s_{LF}(q,d) =  \alpha s_{v}(q,d) + (1-\alpha) s_{t}(q,d).
\end{align}

One weakness of the late fusion strategy is that it combines all gallery items.
Instead, we can limit the influence of the tails to avoid misranking by using
different fusion strategies.
Given $k>0$, let $I_k$ be the indicator function that a gallery item is in the
top-$k$ ranked items according to $s_t$, i.e. $I_k(q,d)=1$ if $d$ is in the
top-k results when querying with $q$ and similarity $s_t$, and $I_k(q,d)=0$
otherwise.
Following~\cite{ah2010leveraging, clinchant2011semantic, csurka2012empirical}, we then define the \textit{late semantic combination (LSC)}
and \textit{product semantic combination (PSC)} with 
Equations~(\ref{eq:LSC}) and~(\ref{eq:PSC}) respectively.
Note that LSC is equivalent to the late fusion if $k$ is equal to the gallery size.
\begin{align}
  s_{LSC}(q,d) &=  \alpha s_{v}(q,d) + (1-\alpha) s_{t}(q,d) I_k(q,d)
  \label{eq:LSC} \\
    s_{PSC}(q,d) &=  s_{v}(q,d) s_{t}(q,d) I_k(q,d)
  \label{eq:PSC}
\end{align}

These different reranking strategies do not require any training and rely on existing
pretrained cross-modal retrieval models. 
We simply use the part of CTC disjoint from the two test sets to choose the hyperparameters $\alpha$ and $k$.

\subsection{STARNet: A Dedicated Trimodal Architecture}
\label{sec:method_sup}

All previously described approaches rely on a pretrained cross-modal retrieval
model. Here, we introduce a new architecture able to handle the trimodality of
the \stacmr{} task. 
We start from the model presented in~\cite{li2019visual} and extend it to
integrate scene text. First, we assume that scene text has been detected within an image.
Then we adapt the model of~\cite{li2019visual} to be able to read scene-text
instances. We include a positional information encoder along with a scene-text Graph Convolutional Network (GCN) and a customized
fusion module into the original pipeline.
Sharing intuition with~\cite{vo2019composing}, we assume that scene text acts as a
modifier in the joint embedding space, applied to the visual descriptor
of an image.

\begin{figure}[t!]
\begin{center}
\includegraphics[width=\linewidth]{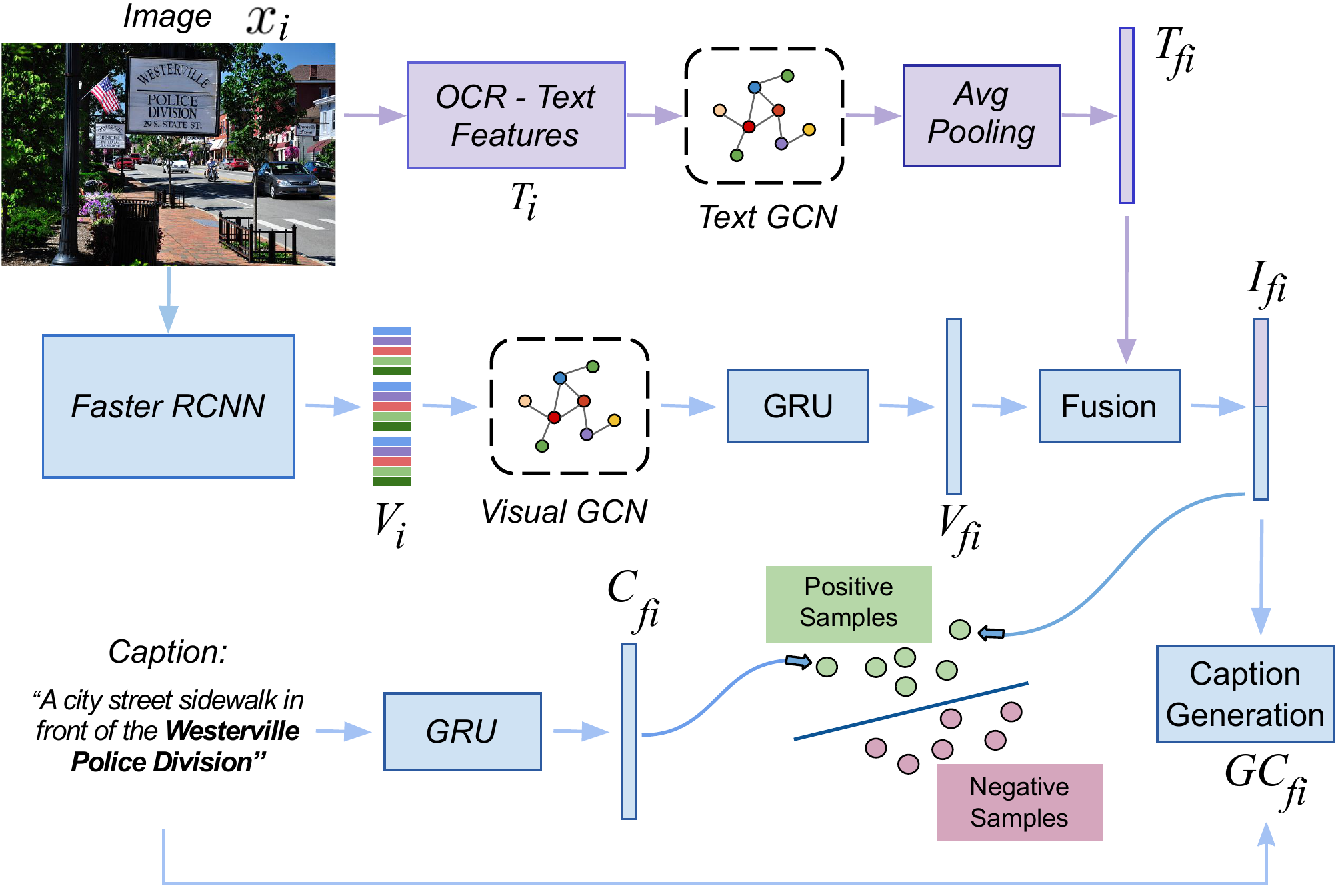}
\end{center}
\caption{\textbf{Our proposed STARNet model.} Visual regions and scene-text instances are used as input to a GCN. The final learned representations are later combined to leverage complementary semantic information.}
\label{fig:supervised_model}
\end{figure}

We propose the STARNet (\textbf{S}cene-\textbf{T}ext \textbf{A}ware
\textbf{R}etrieval \textbf{Net}work) model, illustrated in
Figure~\ref{fig:supervised_model}.
It is composed of the following modules: a joint encoder $\Phi$ for both an image and its
scene text, a caption encoder $\Theta$, and a caption generation module $\Psi$.
Given an image $I_i$ and its scene-text $OCR_i$, the global feature encoding for
both modalities is 
$ I_{fi} = \Phi (I_i, OCR_i)$.
The image encoder follows~\cite{anderson2018bottom} and uses a customized Faster
R-CNN~\cite{ren2015faster} to extract visual features for all regions of
interest represented by $V_i$. Similarly, the employed OCR~\cite{GoogleOCR2020}
extracts scene-text instances as well as their bounding boxes and is represented by $T_i$. 

For both modalities, image and scene text, we use a GCN~\cite{kipf2016semi} to obtain richer representations. For notation purposes we refer to the visual or textual features as $F_{i}$ since the formulation of both visual and textual GCNs are similar. The inputs to each GCN are features 
$F_{fi}\in{R}^{k\times D}$, where $D=2048$ and, $k=36$ in the case of $V_{i}$ and $k=15$ in the case of $T_{i}$. A zero padding scheme is employed for both modalities if the number of features is smaller than $k$.
We define the affinity matrix $R$, which computes the correlation between two regions and is given by: $ R_{ij} =  \rho(k_i)^T  \omega(k_j) $, where $k_i, k_j$ represent the two features being compared and $\rho(.)$ and $\omega(.)$ are two fully connected layers that are learned in an end-to-end manner by back propagation. 

The obtained graph can be defined by $F_{fi} = (F_{i},R) $, in which the nodes are represented by the features $F_{i}$ and the edges are described by the affinity matrix $R$. The graph describes through $R$ the degree of semantic relation between two nodes. In our method, we employ the definition of Graph Convolutional Networks given by~\cite{kipf2016semi} to obtain a richer set of features from the nodes and edges. The equation that describes a single Graph Convolution layer is:
\begin{equation}
F^{l}_g = W_r^{l}(R^{l} F_{i}^{l-1} W_g^{l}) + F_{i}^{l-1}
\end{equation}
where $R \in \mathbb{R}^{k \times k}$ is the affinity matrix , $F_{i} \in \mathbb{R}^{k \times D}$ are the input features of a previous layer, $W_g \in \mathbb{R}^{D \times D}$ is a learnable weights matrix of the GCN, $W_r\in \mathbb{R}^{k \times k}$ is a residual weights matrix and $l$ is the number of GCN layer. Particularly, we employ a total number of $l=4$ for both GCNs used in the proposed pipeline.

The output of the visual GCN goes through a Gated Recurrent Unit (GRU)~\cite{chung2014empirical} to obtain the global image representation denoted by $V_{fi}$. Textual features from the output of the scene-text GCN are average-pooled to obtain a final textual representation denoted by $T_{fi}$.
The final image representation $I_{fi}$ is the dot product between the visual and final scene-text features (which act as a modifier) added to the original visual features: 
$I_{fi} = V_{fi} \odot T_{fi} + V_{fi}$.

Caption $C_i$ from the corresponding training image-caption pair
is encoded with a GRU~\cite{chung2014empirical, faghri18vse}, leading to $C_{fi} = \Theta (C_i)$. 
To align image features with their caption features in a joint embedding space, we train $\Phi$ and $\Theta$ using a triplet ranking loss~\cite{faghri18vse, lee2018stacked} by employing the hardest negative sample on each mini-batch.

In order to provide the model with a stronger supervision signal, the learned image representation $I_{fi}$ is also used to generate a caption as an auxiliary task. We train the third encoder $\Psi$ so that the generated caption equals to: $GC_{fi} = \Psi(I_{fi})$.
This sequence to sequence model uses an attention mechanism similarly to~\cite{venugopalan2015sequence} and we optimize the log-likelihood of the predicted output caption given the final visual features and the previous generated word.


\section{Experiments}
\label{sec:exp}

We present results on CTC. 
They are split into two parts: visual-only and scene-text-only baselines, as well as
their unsupervised re-ranking (Section~\ref{sec:baselines}), and supervised
trimodal fusion results from STARNet (Section~\ref{sec:starnet}).
Following cross-modal retrieval (CMR) evaluation standards, we report performance with recall at K (R@K) for K in $\{1,5,10\}$ for both image-to-text and text-to-image retrieval.

\begin{table*}[!h]
\begin{center}
\footnotesize
\setlength\tabcolsep{2pt}

\begin{tabular}{cc|c|cc|c|c|rrr|rrr|rrr|rrr}
\toprule
    & \multirow{3}{*}{ \textbf{Visual Model} }                   &         \multirow{3}{*}{\begin{tabular}[c]{@{}c@{}} \textbf{Scene-text} \\ \textbf{Model} \end{tabular} }             &    \multicolumn{2}{c|}{\multirow{2}{*}{\textbf{Trained on}}} & 
    \multirow{3}{*}{\begin{tabular}[c]{@{}c@{}} \textbf{Scene-text} \\ \textbf{Source} \end{tabular} } &
    \multirow{3}{*}{ \textbf{Re-rank} }        & \multicolumn{6}{c|}{\textbf{CTC-1K}}                                                                                                                                  & \multicolumn{6}{c}{\textbf{CTC-5K}}                                                                                                                                  \\ \cline{8-19}
      &                 &                  &   &    & &       & \multicolumn{3}{c|}{\textbf{Image to Text}}                                            & \multicolumn{3}{c|}{\textbf{Text to Image}}                                            & \multicolumn{3}{c|}{\textbf{Image to Text}}                                            & \multicolumn{3}{c}{\textbf{Text to Image}}                                            \\ \cline{4-5} \cline{8-19}
       &                   
       &         & F30K    &  TC         &        &  & \multicolumn{1}{l}{R@1} & \multicolumn{1}{l}{R@5} & \multicolumn{1}{l|}{R@10} & \multicolumn{1}{l}{R@1} & \multicolumn{1}{l}{R@5} & \multicolumn{1}{l|}{R@10} & \multicolumn{1}{l}{R@1} & \multicolumn{1}{l}{R@5} & \multicolumn{1}{l|}{R@10} & \multicolumn{1}{l}{R@1} & \multicolumn{1}{l}{R@5} & \multicolumn{1}{l}{R@10} \\
\hline
\hline
(1) & VSE++~\cite{faghri18vse}       & \xmark & \cmark     & \xmark & - & - & 20.5                    & 42.8                    & 54.5                     & 15.4             & 35.2     & 48.4                     & 13.3                   & {30.2}       & {40.2}                    & {8.4}                     & {21.5}       & {30.1}                     \\
(2) & VSE++   & \xmark & \cmark     & \cmark  & - & -   & {23.9}     & {50.6} & {63.2}      & {16.5}  & {39.6} & {53.3}                     & 12.6                    & 30.1                    & {40.2}                   & 7.9                     & 21.0                    & 29.7                     \\
(3) & VSRN~\cite{li2019visual}  &  \xmark & \cmark     & \xmark & - & -  & 27.1                    & 50.7                    & 62.0                     & 19.7                    & 42.8                    & 55.7                     & 19.2                    & 38.6                    & 49.4                     & 12.5                    & 29.2                    & 39.1                     \\
(4) & VSRN  & \xmark & \cmark     & \cmark & -  & -  & \textbf{35.6}        & \textbf{64.4}                    & \textbf{76.0}            & \textbf{24.1}          & \textbf{50.1}                    & \textbf{63.8}         & \textbf{22.7}             & \textbf{45.1}          & \textbf{56.0}       & \textbf{14.2}  & \textbf{32.1}            & \textbf{42.6}                     \\
           \hline \hline
(5) &  \xmark & VSE++  GRU  & \cmark     & \cmark  & GT & -    & \textbf{26.3}        & \textbf{40.4}      & \textbf{47.3}    & \textbf{10.0}    & \textbf{20.3} & \textbf{25.6}  & \textbf{4.4}       & \textbf{7.1}     & \textbf{8.2}       & \textbf{1.6}       & \textbf{3.5}       & \textbf{4.7}                      \\
(6) & \xmark & VSRN GRU & \cmark & \cmark & GT & -              & 12.3      & 25.1      & 30.1      & 6.8       & 15.3      & 20.0      & 1.9       & 4.0       & 5.2       & 1.1       & 2.8       & 3.8       \\
(7) &  \xmark  & Fasttext+FV & \xmark     & \xmark & GT  & - & {21.7}                    & {36.5}                    & {44.3}                     & 3.2                     & 6.6                     & 9.0                      & {3.5}               & {5.9}                     & {7.5}                      & 0.6                     & 1.3                     & 1.7                      \\
    \hline \hline
(8) & \multirow{4}{*}{VSE++} & \multirow{4}{*}{VSE++ GRU} & \multirow{4}{*}{\cmark}     & \multirow{4}{*}{\cmark}  & \multirow{4}{*}{GT}  &
 AVG   &  34.6	& 53.1	& 61.0 & 14.5	& 31.0	& 39.4	& 10.0	& 21.5	& 29.5	& 5.0	& 14.1	& 21.4
    \\
(9) &  &  & & & & LF          &     31.0	& 60.0 &	72.3 & \textbf{20.4}	& \textbf{44.7} & 57.3 & 13.4 & 30.9 & 41.5& 7.4 & 20.5 & 29.1         \\
 (10) &                      &                                  &                                  & &                       & PSC         &       \textbf{37.4} & \textbf{62.8}	& \textbf{73.6} & 15.5 & 42.6	& 57.1	& 12.2	& \textbf{32.1}	& \textbf{42.4}	& 4.1	& 19.3	& 29.2    \\
 (11) &                       &                                                                    & &                  &      & LSC         &      31.6	& 57.8	& 70.2	& 20.3	& \textbf{44.7}	& \textbf{57.8}	& \textbf{13.7}	& 31.7	& 41.6	& \textbf{7.7}	& \textbf{21.0}	& \textbf{29.6} \\
                       \hline
(12) & \multirow{4}{*}{VSRN }  & \multirow{4}{*}{VSRN  GRU} & \multirow{4}{*}{\cmark}     & \multirow{4}{*}{\cmark}  & \multirow{4}{*}{GT}&
 AVG  & 36.8	& 62.2	& 72.9	& 18.6	& 40.5	& 52.9	& 15.3	& 33.5	& 44.3	& 6.4	& 18.9	& 28.0 \\
(13) & & & & & & LF          & \textbf{40.3}           & \textbf{68.5}           & \textbf{79.9}            & 23.9                    & 49.9                    & 63.4                     & 22.6                    & 45.0                    & 56.3                     & 11.8                    & 29.5                    & 40.0                     \\
(14) &                     &                                                 &                   & &                    & PSC         & 33.5                    & 65.9                    & 78.2                     & 15.8                    & 48.1                    & \textbf{64.3}                     & 18.5                    & 44.5                    & 56.0                     & 5.3                     & 28.7                    & 41.0                     \\
(15) &                    &                                                   &                     &    &              & LSC         & 38.6                    & 67.5                    & 78.5                     & \textbf{24.3}           & \textbf{50.4}           & 64.0            & \textbf{23.4}           & \textbf{45.6}           & \textbf{56.5}            & \textbf{12.1}           & \textbf{30.6}           & \textbf{41.1}            \\
                       \hline
(16) & \multirow{3}{*}{VSRN }  & \multirow{3}{*}{VSE++  GRU} & \multirow{3}{*}{\cmark}     & \multirow{3}{*}{\cmark} & \multirow{3}{*}{GT} & LF          & \textbf{45.8}	& \textbf{72.7}	& 81.4	& 26.5	& 52.7	& 66.1 & 24.2 & 46.1 & 57.1 & 12.9 & 31.0 & 41.2                    \\
(17) &                      &                   & &                                                       &                & PSC         & 42.2	& 71.5	& \textbf{82.8}	&	18.9	& 51.1	& \textbf{66.4} & 20.1	& 46.4	& \textbf{57.5}	&	6.7	& 29.5	& 41.6 \\
(18) &                     &                       & &                                                     &              & LSC         & 45.3	& 71.5	& 80.7	& \textbf{26.7}	& \textbf{53.0}	& 66.2	& \textbf{24.4}	& \textbf{46.9}	& 57.4	& \textbf{13.2}	& \textbf{31.8}	& \textbf{42.3} \\
  \hline
  \hline
(19) & \multirow{3}{*}{VSRN }  & \multirow{3}{*}{VSE++  GRU}   & \multirow{3}{*}{\cmark}     & \multirow{3}{*}{\cmark}  & \multirow{3}{*}{OCR}  & LF          &  
 41.5 & \textbf{70.1} & 79.8 & 25.1 & 51.2 & 64.3 & \textbf{23.3} &	45.0 & \textbf{58.9} & 12.6 & 30.5 & 41.1 \\
(20) &                     &                                              & &                                     &       & PSC         & 38.5 & 69.6 & \textbf{80.6} & 17.9 & 50.1 & \textbf{65.1} & 19.8 & \textbf{45.7}	 & 57.2 & 7.0	& 29.8 & 41.7 \\
(21) &                     &                                                 & &                                  &       & LSC         & \textbf{42.2} & 68.6 & 78.5 & \textbf{25.5} & \textbf{51.8} & 64.9 & 19.8	& \textbf{45.7} &  57.2 & \textbf{13.2} & \textbf{31.5} & \textbf{42.2}     \\      
\bottomrule
\end{tabular}
\end{center}
\vspace{-3mm}
\caption{Results on CTC for visual and scene-text baselines, and their re-ranking combinations. 
\textbf{Visual model} and \textbf{Scene-text model} indicate image-caption and scene-text-caption retrieval, respectively. 
\textit{GT} stands for ground-truth scene-text annotations and \textit{OCR} for scene-text prediction obtained from~\cite{GoogleOCR2020}.
\textbf{Bold} numbers denote the best performances of visual, scene-text, and re-ranking methods for each ensemble of models. }
\label{megatable}
\end{table*}

\subsection{Baselines and Re-Ranking Results} 
\label{sec:baselines}
This section first introduces visual-only CMR models. These allow observing how standard CMR models tackle the \stacmr{} task on CTC. Then, we propose scene-text-only metric spaces, where the only information extracted from the image is its scene text. These baselines should help judge the semantic relevance of the scene-text with respect to the captions. 
The remaining results correspond to different combinations: a naive average of visual and scene-text embeddings for metric spaces that allow it, and the different re-ranking strategies introduced in Section~\ref{sec:method_unsup}.

\textbf{Visual-only Baselines.}
We use two CMR models based on global features for both images and captions,
VSE++~\cite{faghri18vse} and VSRN~\cite{li2019visual}. Both works provide public training code, used for all models in this section, with the exception of the VSE++ model trained on Flickr30K, for which we use the model provided by \cite{faghri18vse}. We train these architectures either with Flickr30K or Flickr30K + TextCaps. As mentioned in Section \ref{sec:dataset_collection}, models pretrained on COCO Captions are not considered due to the overlap between the training set of COCO Captions and our test sets.

Results are presented in Table \ref{megatable}, rows (1-4). VSRN surpasses VSE++, mirroring their relative performance from CMR benchmarks. Furthermore, models trained on the additional data of TextCaps outperform models trained only on Flickr30k. This is interesting, as TextCaps images-captions pairs are more dependent on their scene text than those from Flickr30k. 
Enlarging the dataset size with the inclusion of TextCaps explains this improvement to an extent, as the training set of Flickr30k is relatively small. Moving forward, we only report models trained on F30K+TC.

\textbf{Scene-Text only Baselines.}
We use the textual embedding part of our two previously used CMR models 
(denoted by VSE++ GRU and  VSRN GRU respectively).
We also consider FastText~\cite{bojanowski2017enriching} word embeddings followed by a Fisher vector encoding~\cite{perronnin2007fisher} (denoted by FastText+FV), which is able to deal with out-of-vocabulary words. For these experiments, we use the ground-truth OCR annotations as scene text.
Results are presented in Table \ref{megatable}, rows (5-7). We observe much weaker results than the purely visual baselines. For CTC-1K, this approach can rely on shared words between scene text and one of the captions. For the more realistic CTC-5K, we see that scene text brings very little in isolation.
Note that the VSE++ GRU outperforms VSRN GRU for this task, while VSRN is better for the purely visual case. 
This motivates the hybrid strategies merging both models that we report later. 
Fasttext+FV yields strong results on image-to-caption retrieval on CTC-1K, but shows poor results on the other evaluated scenarios. A discussion of several scene-text only baselines is available in the supplementary material.

\textbf{Average Embedding.}
If an image and scene text are represented using the same CMR model, all three modalities are represented in the same embedding space. 
This allows a naive combination which consists in 
averaging visual and scene-text embeddings to represent the image, reported as AVG on the Table \ref{megatable}, rows (8) and (12). 
This brings a non-negligible improvement on CTC-1K Image to Text compared to their respective visual-only baseline
and it is a first proof that scene text, even naively used, improves on some \stacmr{} queries. 
However, we observe
a decline on CTC-5K in the same comparison. 
This hints at the fact that scene text provides fine-grained information that should be used selectively, and giving equal weight to both modalities is too crude an approach.

\textbf{Re-Ranking Results.}
Some re-ranking results are presented in Table \ref{megatable}, rows (9-21). 
We test the best pairing of visual-only and scene-text-only models with three combination strategies: late fusion (LF), product semantic combination (PSC) and late semantic combination (LSC). 
Hyper-parameters of each re-ranking strategy are chosen for VSRN with VSE++ GRU 

and applied to all other combinations as is. 
We use the part of CTC explicit which is not used for testing as validation. 
For LF, $\alpha=0.8$. For PSC, $\alpha=0.95$ and $k=3$. For LSC, $\alpha=0.8$ and $k=100$.

When compared to the unimodal baselines, all combinations improve results on CTC-1K. 
Both LF and LSC  match the results of their visual baselines on CTC-5K, showing that these methods are more robust to scene-text information unrelated to the captions.

For the three best performing re-ranking variants,
we repeat the experiment using OCR predictions instead of the ground-truth scene-text annotations. Results are shown in rows (19-21). When compared with their counterparts in rows (16-18), we observe a R@10 loss on average of 1.7$\%$ in CTC-1k and stable results for CTC-5k. This validates the stability of these re-ranking strategies to loss of information due to imperfect OCR predictions.

\begin{table*}[!ht]
\begin{center}
\footnotesize
\setlength\tabcolsep{2pt}

\begin{tabular}{l|c|c|ccc|ccc|ccc|ccc|ccc}
\toprule

\multirow{3}{*}{\textbf{Model}} & \multirow{3}{*}{\textbf{\begin{tabular}[c]{@{}c@{}}Uses \\ Scene Text\end{tabular}}} & \multirow{3}{*}{\textbf{\begin{tabular}[c]{@{}c@{}}Scene-Text\\ Source\end{tabular}}} & \multicolumn{3}{c|}{\multirow{2}{*}{\textbf{Trained on}}} & \multicolumn{6}{c|}{\textbf{CTC-1K}}                                    & \multicolumn{6}{c}{\textbf{CTC-5K}}                                    \\ \cline{7-18} 
                                &                                                                                      &                                                                                       & \multicolumn{3}{c|}{}       & \multicolumn{3}{c|}{Image to Text} & \multicolumn{3}{c|}{Text to Image} & \multicolumn{3}{c|}{Image to Text} & \multicolumn{3}{c}{Text to Image} \\ \cline{4-18} 
                                &                                                                                      &                                                                                       & F30K                        & TextCaps       & CTC        & R@1        & R@5       & R@10      & R@1        & R@5       & R@10      & R@1        & R@5       & R@10      & R@1        & R@5       & R@10      \\ \hline
\multirow{5}{*}{\textbf{SCAN}~\cite{lee2018stacked}}           & \xmark                                                                                & -                                                                                     & \cmark                       & \xmark          & \xmark      & 26.4       & 48.6      & 61.1      & 15.2       & 36.8      & 49.3      & 17.5       & 36.7      & 47.1      & 7.6        & 21.2      & 30.4      \\
                                & \cmark                                                                                & OCR                                                                                   & \xmark                       & \cmark          & \xmark      & 19.5       & 43.8      & 57.1      & 10.2       & 28.7      & 42.1      & 7.0        & 20.0      & 29.7      & 3.2        & 11.7      & 18.1      \\
                                & \cmark                                                                                & OCR                                                                                   & \cmark                       & \cmark          & \xmark      & 35.0       & 62.9      & 74.4      & 19.3       & 44.0      & 58.3      & 21.1       & 43.0      & 54.6      & 9.6        & 25.4      & 35.6      \\
                                & \cmark                                                                                & OCR                                                                                   & \cmark                       & \xmark          & \cmark      & 27.5       & 48.9      & 61.9      & 16.5       & 37.7      & 51.1      & 18.6       & 37.3      & 47.6      & 8.1        & 21.6      & 30.6      \\
                                & \cmark                                                                                & OCR                                                                                   & \cmark                       & \cmark          & \cmark      & 36.3       & 63.7      & 75.2      & 26.6       & 53.6      & 65.3      & 22.8       & 45.6      & 54.3      & 12.3       & 28.6      & 39.9      \\ \hline
\multirow{5}{*}{\textbf{VSRN}~\cite{li2019visual}}           & \xmark                                                                                & -                                                                                     & \cmark                       & \xmark          & \xmark      & 27.1                    & 50.7                    & 62.0                     & 19.7                    & 42.8                    & 55.7                     & 19.2                    & 38.6                    & 49.4                     & 12.5                    & 29.2                    & 39.1      \\
                                & \cmark                                                                                & OCR                                                                                   & \xmark                       & \cmark          & \xmark      & 18.6       & 40.4      & 52.2      & 11.7       & 31.0      & 44.2      & 6.6        & 17.9      & 25.8      & 4.5        & 13.0      & 19.8      \\
                                & \cmark                                                                                & OCR                                                                                   & \cmark                       & \cmark          & \xmark      & 35.6       & 64.3      & 76.0      & 24.0       & 50.1      & 63.1      & 22.6       & 45.0      & 55.9      & 14.2       & 32.1      & 42.5      \\
                                & \cmark                                                                                & OCR                                                                                   & \cmark                       & \xmark          & \cmark      & 36.1       & 64.1      & 75.8      & 26.2       & 53.1      & 65.2      & 24.6       & 48.1      & 58.8      & 15.4       & 35.7      & 46.9      \\
                                & \cmark                                                                                & OCR                                                                                   & \cmark                       & \cmark          & \cmark      & 38.2       & 67.4      & 79.1      & 26.6       & 54.2      & 66.2      & 23.7       & 47.6      & 59.1      & 14.9       & 34.7      & 45.5      \\ \hline
\multirow{5}{*}{\textbf{STARNet}}        & \xmark                                                                                & OCR                                                                                   & \cmark                       & \xmark          & \xmark      & 29.4       & 52.3      & 62.6      & 21.8       & 44.3      & 57.2      & 19.9       & 39.6      & 50.1      & 13.4       & 30.7      & 40.4      \\
                                & \cmark                                                                                & OCR                                                                                   & \xmark                       & \cmark          & \xmark      & 23.4       & 48.0      & 61.0      & 14.2       & 34.9      & 47.3      & 5.1        & 15.1      & 22.3      & 3.9        & 11.9      & 25.1      \\
                                & \cmark                                                                                & OCR                                                                                   & \cmark                       & \cmark          & \xmark      & 39.3       & 65.4      & 76.8      & 25.9       & 52.3      & 65.2      & 21.1       & 41.8      & 52.9      & 13.8       & 31.8      & 42.0      \\
                                & \cmark                                                                                & OCR                                                                                   & \cmark                       & \xmark          & \cmark      & 36.5       & 64.6      & 74.3      & 26.4       & 53.8      & 65.6      & 25.5       & 48.4      & 59.8      & 15.7       & 35.3      & 46.6      \\
                                & \cmark                                                                                & OCR                                                                                   & \cmark                       & \cmark          & \cmark      & \textbf{44.1}       & \textbf{74.8}      & \textbf{82.7}      & \textbf{31.5}       & \textbf{60.8}      & \textbf{72.4}      & \textbf{26.4}       & \textbf{51.1}      & \textbf{63.9}      & \textbf{17.1}       & \textbf{37.4}      & \textbf{48.3}      \\ \hline
\textbf{Re-rank Comb. (21)}               & \cmark                                                                                & OCR                                                                                   & \cmark                       & \cmark          & \xmark      & 42.2       & 68.6      & 78.5      & 25.5       & 51.8      & 64.9      & 19.8       & 45.7      & 57.2      & 13.2       & 31.5      & 42.2      \\\hline \hline
\textbf{STARNet - GT}                         & \cmark                                                                                & GT                                                                                    & \cmark                       & \cmark          & \cmark      & 45.4       & 74.9      & 83.9      & 32.0       & 61.2      & 73.3      & 26.8       & 51.4      & 64.1      & 17.4       & 37.8      & 48.7\\      
\bottomrule
\end{tabular}

\end{center}
\vspace{-3mm}
\caption{Retrieval results on the CTC-1K and CTC-5K test set of \textbf{supervised} models. Second-to-last row shows the result from the unsupervised re-ranking baseline described in Table~\ref{megatable}, line 21.
\textit{OCR} stands for the textual features obtained from~\cite{GoogleOCR2020}, whereas \textit{GT} refers to the Ground-truth annotated scene text. Results depicted in terms of Recall@K (R@K).
}

\label{tab:supervised_full_results}
\end{table*}
\subsection{Supervised Results}\label{sec:starnet}
Latest cross-modal retrieval models rely on region-based visual features%
~\cite{lee2018stacked, li2019visual, wang2020consensus} rather than a global image representation~\cite{faghri18vse}. 
In this section, we include results of 
two state-of-the-art models, SCAN~\cite{lee2018stacked} and VSRN~\cite{li2019visual} that employ such region-based visual features. 
The original cross-modal retrieval models, SCAN and VSRN are used only when trained on Flickr30K. In order to leverage scene text, we have modified them to include OCR features. In both models, the OCR features are projected into the same space as the visual features and the default hyperparameters are employed, details are showed in the supplementary material.
All the obtained results are reported on Table~\ref{tab:supervised_full_results}. 
The second column depicts the usage of scene-text instances by each model, and the third column depicts the source of the scene text. 
We make the following observations.

First, 
we see that using standard models trained on a common cross-modal retrieval dataset, such as Flickr30k, does not yield good performances on the \stacmr{} task. 

Second, we note the different behaviors when each dataset is used for training and testing is done on the CTC test sets. In particular, it is worth noting that by training solely on TextCaps~\cite{sidorov2020textcaps}, the performance of any model decreases significantly, specially in the CTC-5K dataset. This effect is caused by the bias in Textcaps that places a big focus on scene-text instances to describe an image, rather than combining visual and textual features in an unbiased way.

However, all datasets provide complementary statistics to train the STARNet model. For instance, Flickr30k focuses on relevant visual regions, whereas the combination of TextCaps and CTC can be seen as a reciprocal set of datasets that aim towards modeling the relevance of scene-text from an image in a more natural manner. 

It is worth pointing out that STARNet almost doubles the performance in the CTC-1K subset when compared to common retrieval models. We believe this effect is due to the explicit scene-text instances that reinforce the notion of the relevance of this modality. A smaller improvement is achieved in the CTC-5K. This result is caused by the fact that even though scene text does not appear explicitly in the captions, a varying degree of semantics between image and scene text can still be found. 

Finally, we also show an upper-bound at test time assuming a perfect OCR (using ground truth scene-text annotations in CTC), which adds a slight boost to the proposed method. This effect shows and confirms the importance of accurate scene-text recognizers in the \stacmr{} task. 
Additional experiments regarding the performance of the baseline supervised models have been conducted in Flickr30K and TextCaps datasets along with qualitative
results available on the supplementary material.

\section{Conclusion}
\label{sec:ccl}
In this work, we highlight the challenges stemming from including scene-text information in the cross-modal retrieval task. Although of high semantic value, scene text proves to be a fine-grained element in the retrieval process that should be used selectively. We introduce a realistic dataset, 
\textit{CTC}, where annotations for both scene text and captions are available. Contrary to datasets constructed with scene text in mind, \textit{CTC} is unbiased in terms of scene-text content and of how  
it is employed in the captions.
A comprehensive set of baseline methods showcase that 
combining modalities is beneficial, while 
a simple fusion 
cannot tackle the newly introduced task of scene-text aware cross-modal retrieval. 
Finally, we introduce \textit{STARNet} a supervised model that successfully combines all three modalities.


{\small
\bibliographystyle{ieee_fullname}
\bibliography{egbib}
}

\appendix

\section{Additions to Baselines and Re-Ranking}

\subsection{Full Table of Results on CTC}
 Table~\ref{megatable_v2} presents a more extensive version of the results presented in Table~\ref{megatable}. This section dives into some parts of these results.
\begin{table*}[!h]
\begin{center}
\scriptsize
\setlength\tabcolsep{2pt}

\begin{tabular}{cc|c|cc|c|c|rrr|rrr|rrr|rrr}
\toprule
    & \multirow{3}{*}{ \textbf{Visual Model} }                   &         \multirow{3}{*}{\begin{tabular}[c]{@{}c@{}} \textbf{Scene-text} \\ \textbf{Model} \end{tabular} }             &    \multicolumn{2}{c|}{\multirow{2}{*}{\textbf{Trained on}}} & 
    \multirow{3}{*}{\begin{tabular}[c]{@{}c@{}} \textbf{Scene-text} \\ \textbf{Source} \end{tabular} } &
    \multirow{3}{*}{ \textbf{Re-rank} }        & \multicolumn{6}{c|}{\textbf{CTC-1K}}                                                                                                                                  & \multicolumn{6}{c}{\textbf{CTC-5K}}                                                                                                                                  \\ \cline{8-19}
      &                 &                  &   &    & &       & \multicolumn{3}{c|}{\textbf{Image to Text}}                                            & \multicolumn{3}{c|}{\textbf{Text to Image}}                                            & \multicolumn{3}{c|}{\textbf{Image to Text}}                                            & \multicolumn{3}{c}{\textbf{Text to Image}}                                            \\ \cline{4-5} \cline{8-19}
       &                   
       &         & F30K    &  TC         &        &  & \multicolumn{1}{l}{R@1} & \multicolumn{1}{l}{R@5} & \multicolumn{1}{l|}{R@10} & \multicolumn{1}{l}{R@1} & \multicolumn{1}{l}{R@5} & \multicolumn{1}{l|}{R@10} & \multicolumn{1}{l}{R@1} & \multicolumn{1}{l}{R@5} & \multicolumn{1}{l|}{R@10} & \multicolumn{1}{l}{R@1} & \multicolumn{1}{l}{R@5} & \multicolumn{1}{l}{R@10} \\
\hline
\hline
(1) & VSE++       & \xmark & \cmark     & \xmark & - & - & 20.5                    & 42.8                    & 54.5                     & 15.4             & 35.2     & 48.4                     & \underline{13.3}                    & {30.2}       & {40.2}                    & {8.4}                     & {21.5}       & {30.1}                     \\
(2) & VSE++   & \xmark & \cmark     & \cmark  & - & -   & \underline{23.9}     & \underline{50.6} & {63.2}      & {16.5}  & {39.6} & {53.3}                     & 12.6                    & 30.1                    & {40.2}                   & 7.9                     & 21.0                    & 29.7                     \\
(3) & VSRN  &  \xmark & \cmark     & \xmark & - & -  & 27.1                    & 50.7                    & 62.0                     & 19.7                    & 42.8                    & 55.7                     & 19.2                    & 38.6                    & 49.4                     & 12.5                    & 29.2                    & 39.1                     \\
(4) & VSRN  & \xmark & \cmark     & \cmark  & - & -  & \textbf{35.6}        & \textbf{64.4}                    & \textbf{76.0}            & \textbf{24.1}          & \textbf{50.1}                    & \textbf{63.8}         & \textbf{22.7}             & \textbf{45.1}          & \textbf{56.0}       & \textbf{14.2}  & \textbf{32.1}            & \textbf{42.6}                     \\
           \hline
           \hline
(5) &  \xmark & VSE++ GRU & \cmark     & \xmark & GT & -  & 17.4                    & 29.9                    & 37.1                     & {8.3}                     & {17.5}                    & {23.2}                     & 2.4                     & 4.8                     & 5.8                      & {1.3}     & {3.0}                     & {4.2}                      \\
(5') &  \xmark & VSE++ GRU & \cmark     & \xmark & OCR & -  & 12.4	& 21.7	& 26.0	& 6.5	& 14.5	& 18.9	& 1.9	& 3.6	& 4.4	& 1.1	& 2.6	& 3.6          \\
(6) &  \xmark & VSE++  GRU  & \cmark     & \cmark & GT   & -    & \textbf{26.3}        & \textbf{40.4}      & \textbf{47.3}    & \textbf{10.0}    & \textbf{20.3} & \textbf{25.6}  & \textbf{4.4}       & \textbf{7.1}     & \textbf{8.2}       & \textbf{1.6}       & \textbf{3.5}       & \textbf{4.7}                      \\
(6') &  \xmark & VSE++  GRU  & \cmark     & \cmark & OCR   & -    & 19.9	& 30.8	& 36.4	& 8.8	& 16.1	& 20.8	& 3.4	& 5.4	& 6.3	& 1.5	& 3.0	& 4.0          \\
(7) & \xmark & VSRN GRU  & \cmark & \xmark & GT  & -             & 7.7       & 18.8      & 26.0      & 5.2       & 12.7      & 18.8      & 1.1       & 2.4       & 3.3       & 0.9       & 2.2       & 3.3       \\
(8) & \xmark & VSRN GRU & \cmark & \cmark & GT  & -              & 12.3      & 25.1      & 30.1      & 6.8       & 15.3      & 20.0      & 1.9       & 4.0       & 5.2       & 1.1       & 2.8       & 3.8       \\
(9) & \xmark & GRU++ & \cmark & \xmark & GT  & -   & 16.0      & 29.9      & 35.1      & 8.7       & 17.7      & 22.4      & 1.4       & 2.5       & 3.5       & 0.8       & 2.0       & 2.9       \\
(10) & \xmark & Fasttext+FV uncleaned & \xmark & \xmark & GT  & -    & 19.5      & 35.8      & 43.1      & 0.5       & 1.4       & 2.1       & 3.1       & 5.4       & 7.1       & 0.1       & 0.3       & 0.4       \\

 (11) &  \xmark  & Fasttext+FV & \xmark     & \xmark  & GT  & - & {21.7}                    & {36.5}                    & {44.3}                     & 3.2                     & 6.6                     & 9.0                      & {3.5}               & {5.9}                     & {7.5}                      & 0.6                     & 1.3                     & 1.7                      \\
    \hline
    \hline
(12) & \multirow{4}{*}{VSE++} & \multirow{4}{*}{VSE++ GRU} & \multirow{4}{*}{\cmark}     & \multirow{4}{*}{\xmark}  & \multirow{4}{*}{GT}  &
 AVG   & 31.1                    & 54.5                    & 65.7                     & 17.2                    & 37.2                    & 47.6                     & 7.2                     & 16.4                    & 24.0                     & 4.7                     & 13.5                    & 20.7    
    \\
(13) &  &  & & & & LF          & 25.3                    & 51.9           & 63.6            & 17.3           & 39.5           & 52.2                     & 13.4                    & 30.1                    & 40.4                     & 7.5                     & 20.3                    & 29.2                     \\
 (14) &                      &                                  &                                  & &                       & PSC         & 25.8                    & 51.7                    & 63.2                     & 13.5                    & 37.4                    & 51.0                     & 10.9                    & 30.5                    & 41.3                     & 4.2                     & 19.8                    & 29.5                     \\
 (15) &                       &                                                                    & &                  &      & LSC         & 25.9           & 51.8                    & 63.1                     & 17.2                    & 39.4                    & 52.5            & 13.6           & 31.1           & 41.5            & 7.9            & 20.8           & 30.0            \\
                       \hline
(16) & \multirow{3}{*}{VSRN}  & \multirow{3}{*}{VSE++ GRU } & \multirow{3}{*}{\cmark}     & \multirow{3}{*}{\xmark}  & \multirow{3}{*}{GT} & LF          &   35.6	& {61.2} & {71.3}	& 21.8	& 45.4	& 58.0	& 19.2	& 39.2	& 50.2	& 10.7	& 26.7	& 36.9                   \\
(17) &                      &                                          &                                 & &                 & PSC         & 30.6    & 59.3    & 69.5    & 16.2    & 43.2    & {58.2}    & 14.8    & 38.8    & 50.2    & 6.0    & 26.4    & 38.1 \\
(18) &                     &                                              &                            & &              & LSC         & {38.0} & 60.3 & 70.3 & {21.9} & {45.8} & {58.2} & {20.3} & {40.0} & {50.6} & 11.1 & 27.8 & 38.2 \\
                       \hline                
(19) & \multirow{3}{*}{VSRN}  & \multirow{3}{*}{VSE++ GRU } & \multirow{3}{*}{\cmark}     & \multirow{3}{*}{\xmark}  & \multirow{3}{*}{OCR} & LF          &                   32.2 & 58.3 & 69.3 & 20.3 & 43.5 & 56.5 & 18.3 & 37.8 & 48.5 & 10.6 & 27.0 & 36.8              \\
(20) &                      &                                            &                               & &                 & PSC         & 26.7 & 56.0 & 66.7 & 15.0 & 44.2 & 57.4 & 14.5 & 38.1 & 49.5 & 6.2 & 26.4 & 38.0 \\
(21) &                     &                                                &                          & &              & LSC         & 32.8 & 57.0 & 68.5 & 20.7 & 44.0 & 57.1 & 19.7 & 39.6 & 50.3 & {11.3} & {27.9} & {38.3} \\
                       \hline 
(22) & \multirow{4}{*}{VSE++} & \multirow{4}{*}{VSE++ GRU} & \multirow{4}{*}{\cmark}     & \multirow{4}{*}{\cmark}  & \multirow{4}{*}{GT}  &
 AVG   &  34.6	& 53.1	& 61.0 & 14.5	& 31.0	& 39.4	& 10.0	& 21.5	& 29.5	& 5.0	& 14.1	& 21.4
    \\
(23) &  &  & & & & LF          &     31.0	& 60.0 &	72.3 & 20.4	& 44.7 & 57.3 & 13.4 & 30.9 & 41.5& 7.4 & 20.5 & 29.1         \\
 (24) &                      &                                  &                                  & &                       & PSC         &       37.4 & 62.8	& 73.6 & 15.5 & 42.6	& 57.1	& 12.2	& 32.1	& 42.4	& 4.1	& 19.3	& 29.2    \\
 (25) &                       &                                                                    & &                  &      & LSC         &      31.6	& 57.8	& 70.2	& 20.3	& 44.7	& 57.8	& 13.7	& 31.7	& 41.6	& 7.7	& 21.0	& 29.6 \\
                       \hline
(26) & \multirow{4}{*}{VSRN }  & \multirow{4}{*}{VSRN  GRU} & \multirow{4}{*}{\cmark}     & \multirow{4}{*}{\cmark}  & \multirow{4}{*}{GT} &
 AVG  & 36.8	& 62.2	& 72.9	& 18.6	& 40.5	& 52.9	& 15.3	& 33.5	& 44.3	& 6.4	& 18.9	& 28.0 \\
(27) &  &  & & & & LF          & 40.3           & 68.5           & 79.9            & 23.9                    & 49.9                    & 63.4                     & 22.6                    & 45.0                    & 56.3                     & 11.8                    & 29.5                    & 40.0                     \\
(28) &                     &                                     &                                 & &                    & PSC         & 33.5                    & 65.9                    & 78.2                     & 15.8                    & 48.1                    & 64.3                     & 18.5                    & 44.5                    & 56.0                     & 5.3                     & 28.7                    & 41.0                     \\
(29) &                    &                                         &                                 &    &              & LSC         & 38.6                    & 67.5                    & 78.5                     & 24.3           & 50.4           & 64.0            & 23.4           & 45.6           & 56.5            & 12.1           & 30.6           & 41.1            \\
                       \hline
(30) & \multirow{4}{*}{VSRN }  &   \multirow{4}{*}{VSE++ GRU } & \multirow{4}{*}{\begin{tabular}[c]{@{}l@{}}\cmark \\ \cmark \end{tabular}}     & \multirow{4}{*}{\begin{tabular}[c]{@{}l@{}}\cmark \\ \xmark \end{tabular}}  & \multirow{4}{*}{GT} & LF          & 41.7                    & 68.6           & 78.9            & 25.1           & 52.0           & 65.5            & 22.5                    & 44.4                    & 55.7                     & 12.8                    & 31.0                    & 41.3                     \\
(31) &                      &                                          &                                 & &                 & PSC         & 32.8                    & 67.3                    & 79.9            & 17.6                    & 49.4                    & 64.9                     & 16.1                    & 44.6                    & 56.2            & 6.5                     & 29.3                    & 41.3                     \\
(32) &                     &                                              &                                 & &             & LSC         & 42.2           & 67.9                    & 78.5                     & 25.5           & 52.0           & 65.6            & 23.1           & 45.9           & 56.1                     & 13.3           & 31.7           & 42.2            \\ \cline{7-19}
(33) &                     &                                                 &                                 & &          & Oracle LF   & $^\dagger$63.2           & $^\dagger$82.9           & $^\dagger$89.3            & $^\dagger$37.9           & $^\dagger$64.3           & $^\dagger$75.5            & $^\dagger$31.0           & $^\dagger$53.9           & $^\dagger$64.5            & $^\dagger$19.7           & $^\dagger$39.3           & $^\dagger$49.6            \\
                       \hline
(34) & \multirow{3}{*}{VSRN }  & \multirow{3}{*}{VSE++ GRU}     & \multirow{3}{*}{\begin{tabular}[c]{@{}l@{}}\cmark \\ \cmark \end{tabular}}     & \multirow{3}{*}{\begin{tabular}[c]{@{}l@{}}\cmark \\ \xmark \end{tabular}} & \multirow{3}{*}{OCR} & LF          & 39.1                    & 66.7                    & 79.1            & 24.1                    & 50.3                    & 64.3                     & 21.2                    & 43.8                    & 55.4                     & 12.8                    & {31.8}           & 43.0            \\
(35) &                     &                                                 &        & &                                   & PSC         & 31.6                    & 65.2                    & 78.5                     & 16.6                    & 48.6                    & 64.6                     & 15.8                    & 43.9                    & 55.8                     & 6.7                     & 29.4                    & 41.4                     \\
(36) &                     &                                                  &          & &                                & LSC         & 39.3          & 67.4           & 78.7                     & 24.7           & 50.9           & 64.6            & 22.7           & 45.3           & 56.3            & 13.3           & 31.6                    & 42.2         \\      
\hline
(37) & \multirow{4}{*}{VSRN }  & \multirow{4}{*}{VSE++  GRU} & \multirow{4}{*}{\cmark}     & \multirow{4}{*}{\cmark} & \multirow{4}{*}{GT} & LF          & \textbf{45.8}	& \textbf{72.7}	& 81.4	& 26.5	& 52.7	& 66.1 & 24.2 & 46.1 & 57.1 & 12.9 & 31.0 & 41.2                    \\
(38) &                      &                   & &                              &                                           & PSC         & 42.2	& 71.5	& \textbf{82.8}	&	18.9	& 51.1	& \textbf{66.4} & 20.1	& 46.4	& \textbf{57.5}	&	6.7	& 29.5	& 41.6 \\
(39) &                     &                       & &                               &                                      & LSC         & 45.3	& 71.5	& 80.7	& \textbf{26.7}	& \textbf{53.0}	& 66.2	& \textbf{24.4}	& \textbf{46.9}	& 57.4	& \textbf{13.2}	& \textbf{31.8}	& \textbf{42.3} \\
\cline{7-19}
(40) &                     &                          & &                              &                                    & Oracle LF   & $^\dagger$67.9	& $^\dagger$84.8 & $^\dagger$91.1 & $^\dagger$39.2 & $^\dagger$64.8 & $^\dagger$76.2 & $^\dagger$32.9 & $^\dagger$55.3 & $^\dagger$65.2 & $^\dagger$20.1 & $^\dagger$39.7 & $^\dagger$50.3 \\
                       \hline
(41) & \multirow{3}{*}{VSRN }  & \multirow{3}{*}{VSE++  GRU}   & \multirow{3}{*}{\cmark}     & \multirow{3}{*}{\cmark}    & \multirow{3}{*}{OCR} & LF          &  
 41.5 & \underline{70.1} & 79.8 & 25.1 & 51.2 & 64.3 & \underline{23.3} &	45.0 & \underline{58.9} & 12.6 & 30.5 & 41.1 \\
(42) &                     &                                              & &             &                                 & PSC         & 38.5 & 69.6 & \underline{80.6} & 17.9 & 50.1 & \underline{65.1} & 19.8 & \underline{45.7}	 & 57.2 & 7.0	& 29.8 & 41.7 \\
(43) &                     &                                                 & &             &                              & LSC         & \underline{42.2} & 68.6 & 78.5 & \underline{25.5} & \underline{51.8} & 64.9 & 19.8	& \underline{45.7} &  57.2 & \underline{13.2} & \underline{31.5} & \underline{42.2}     \\      
\bottomrule
\end{tabular}
\end{center}
\caption{Results on CTC-1k and CTC-5k for visual-only baselines, scene-text-only baselines and re-ranking combinations of these baselines. \textbf{Bold results} denote the best performance at each of visual model, scene-text model and re-ranking methods. $\dagger$ denotes theoretical upper-bounds to the linear combination re-rankings. (see Section \ref{sec:oracle})}
\label{megatable_v2}
\end{table*}
\begin{table*}[!h]
\begin{center}
\footnotesize
\setlength\tabcolsep{2pt}

\begin{tabular}{cc|c|cc|c|c|rrr|rrr}
\toprule
    & \multirow{3}{*}{ \textbf{Visual Model} }                   &                                   \multirow{3}{*}{  \textbf{Scene-Text Model} }                                                     &    \multicolumn{2}{c|}{\multirow{2}{*}{\textbf{Trained on}}} & 
    \multirow{3}{*}{ \textbf{Scene-text Source} } &
    \multirow{3}{*}{ \textbf{Re-rank} }        & \multicolumn{6}{c}{\textbf{TextCaps}}        \\ \cline{8-13}
      &                 &                                                           &                   &    & &       & \multicolumn{3}{c|}{\textbf{Image to Text}}                                            & \multicolumn{3}{c}{\textbf{Text to Image}}                           \\ \cline{4-5} \cline{8-13}
       &                &   
       &     F30K    &  TC          &                     &  & \multicolumn{1}{l}{R@1} & \multicolumn{1}{l}{R@5} & \multicolumn{1}{l|}{R@10} & \multicolumn{1}{l}{R@1} & \multicolumn{1}{l}{R@5} & \multicolumn{1}{l}{R@10}  \\
\hline
(1) & VSE++ & \xmark & \cmark & \xmark & - & - & 
5.6 &	15.1	& 21.5 & 4.1	& 11.1	& 16.6 \\
(2) & VSRN & \xmark & \cmark & \xmark & - & - &
6.2	& 14.5	& 20.2 & 4.5 & 11.7 & 16.6  \\
(3) & VSE++ & \xmark & \xmark & \cmark & - & - &
\textbf{14.7}	& \textbf{30.9}	& \textbf{40.4}	& \textbf{10.0}	& \textbf{24.3}	& \textbf{32.9} \\
\hline
(4) & \xmark & VSE++ GRU &  \cmark & \xmark & \multirow{2}{*}{GT} & - &
11.5	& 18.7	& 22.0	& 10.3	& 17.5	& 20.1 \\
(5) & \xmark & VSE++ GRU &  \xmark & \cmark &  & - &
\textbf{34.6}	& \textbf{45.7}	& \textbf{49.7}	& \textbf{25.1}	& \textbf{35.0}	& \textbf{37.9} \\
\hline
(6) & \multirow{5}{*}{VSE++ }  & \multirow{5}{*}{VSE++ GRU}   & \multirow{5}{*}{\xmark}     & \multirow{5}{*}{\cmark} & \multirow{5}{*}{GT} & AVG & 
\textbf{42.8}	& \textbf{56.6}	& 62.8	& \textbf{30.8}	& \textbf{46.2}	& \textbf{52.7} \\
(7) &        &           &         &        &   & LF &
33.5	& 54.7	& 63.7	& 22.6	& 40.8	& 50.2 \\
(8) &        &           &         &        &   & PSC &
40.0	& 56.3	& \textbf{64.6}	& 24.7	& 42.3	& 50.7 \\   
(9) &        &           &         &        &   & LSC &
25.7	& 46.0	& 56.1	& 18.0	& 36.0	& 45.3 \\ \cline{7-13}
(10) &        &           &         &        &   & Oracle LF &
 $^\dagger$57.3 & $^\dagger$72.3 & $^\dagger$78.0 & $^\dagger$39.6 & $^\dagger$55.9 & $^\dagger$63.0 \\
\bottomrule
\end{tabular}
\end{center}
\vspace{-3mm}
\caption{Results on TextCaps (validation set) for visual-only baselines, scene-text-only baselines and re-ranking combinations of these baselines. \textit{GT} stands for ground-truth scene-text annotations, which for TextCaps are OCR predictions from~\cite{borisyuk2018rosetta}.
$\dagger$~denotes theoretical upper-bounds to the linear combination re-rankings. (see Section \ref{sec:oracle})}
\label{tab:textcaps}
\end{table*}

\begin{table*}[!h]
\begin{center}
\footnotesize
\setlength\tabcolsep{2pt}

\begin{tabular}{c|ccc|ccc|ccc|ccc|ccc}
\toprule
\multirow{3}{*}{\textbf{Model}} & \multicolumn{3}{c|}{\multirow{2}{*}{\textbf{Trained on}}} & \multicolumn{6}{c|}{\textbf{Flickr30K}}                                                       & \multicolumn{6}{c}{\textbf{TextCaps}}                                                        \\ \cline{5-16} 
                                & \multicolumn{3}{c|}{}       & \multicolumn{3}{c|}{Image to Text}            & \multicolumn{3}{c|}{Text to Image}            & \multicolumn{3}{c|}{Image to Text}            & \multicolumn{3}{c}{Text to Image}            \\ \cline{2-16} 
                                & F30K                        & TextCaps       & CTC        & R@1           & R@5           & R@10          & R@1           & R@5           & R@10          & R@1           & R@5           & R@10          & R@1           & R@5           & R@10          \\ \hline
\multirow{5}{*}{\textbf{SCAN}}           & \cmark                       & \xmark          & \xmark      & 57.2          & 84.4          & 90.5          & 38.6          & 68.4          & 79.1          & 9.3           & 21.7          & 29.8          & 4.7           & 14.1          & 21.2          \\
                                & \xmark                       & \cmark          & \xmark      & 14.1          & 34.6          & 45.0          & 7.8           & 22.7          & 32.1          & 23.2          & 50.5          & 63.5          & 14.1          & 37.6          & 52.1          \\
                                & \cmark                       & \cmark          & \xmark      & 57.6          & 85.3          & 92.4          & 39.2          & 70.0          & 80.2          & 16.6          & 36.6          & 48.7          & 9.3           & 25.4          & 36.4          \\
                                & \cmark                       & \xmark          & \cmark      & 58.1          & 83.2          & 91.5          & 39.6          & 69.8          & 81.3          & 4.4           & 11.2          & 16.2          & 2.4           & 7.2           & 11.3          \\
                                & \cmark                       & \cmark          & \cmark      & 55.1          & 79.6          & 87.1          & 35.5          & 67.2          & 77.3          & 15.4          & 35.2          & 46.9          & 13.4          & 37.1          & 51.8          \\ \hline
\multirow{5}{*}{\textbf{VSRN}}           & \cmark                       & \xmark          & \xmark      & 63.1          & 86.5          & 92.1          & 47.1          & 75.3          & 83.8          & 6.3           & 14.9          & 21.4          & 4.2           & 11.4          & 16.6          \\
                                & \xmark                       & \cmark          & \xmark      & 11.7          & 30.1          & 40.2          & 9.2           & 23.7          & 32.8          & 14.3          & 34.9          & 46.2          & 9.53          & 26.2          & 37.2          \\
                                & \cmark                       & \cmark          & \xmark      & 62.5          & 86.1          & 92.3          & 48.1          & 76.8          & 84.3          & 19.6          & 41.9          & 53.1          & 13.9          & 32.8          & 43.8          \\
                                & \cmark                       & \xmark          & \cmark      & 64.9 & 88.0 & 93.2 & 49.0          & 76.9          & 84.9          & 8.21          & 18.6          & 25.4          & 5.56          & 14.0          & 19.5          \\
                                & \cmark                       & \cmark          & \cmark      & 60.7          & 85.2          & 90.4          & 45.7          & 73.9          & 81.8          & 18.7          & 38.6          & 50.1          & 12.4          & 30.0          & 41.2          \\ \hline
\multirow{5}{*}{\textbf{STARNet}}        & \cmark                       & \xmark          & \xmark      & 63.9          & 86.9          & 92.4          & 48.6          & 76.7          & 84.7          & 6.79          & 15.5          & 21.6          & 4.6           & 12.1          & 17.5          \\
                                & \xmark                       & \cmark          & \xmark      & 13.3          & 29.6          & 39.6          & 9.8           & 24.5          & 34.1          & 28.7 & 53.7 & 65.1 & 19.8 & 40.1 & 51.6 \\
                                & \cmark                       & \cmark          & \xmark      & 62.4          & 85.8          & 92.1          & 47.1          & 76.1          & 84.1          & 24.0          & 48.9          & 60.7          & 17.3          & 37.9          & 49.8          \\
                                & \cmark                       & \xmark          & \cmark      & 63.2          & 87.2          & 92.5          & 49.5          & 78.1          & 85.2          & 7.5           & 17.5          & 25.1          & 5.2           & 13.6          & 19.5          \\
                                & \cmark                       & \cmark          & \cmark      & \textbf{67.5}          & \textbf{88.1}          & \textbf{93.6}          & \textbf{50.7} & \textbf{78.0} & \textbf{85.4} & \textbf{29.5}          & \textbf{53.8}          & \textbf{65.3}          & \textbf{20.8}         & \textbf{42.9}          & \textbf{53.6}  \\
                                \bottomrule
\end{tabular}

\end{center}
\caption{Quantitative comparison of experimental results of image-to-text and text-to-image retrieval on the Flickr30K (test) and TextCaps (val) sets of supervised models. Metric depicted in terms of Recall@K (R@K). }

\label{tab:supervised_full_results_flickr_ctc}
\end{table*}

\textbf{Scene-Text-only Baselines.}
Here we discuss additional scene-text baselines we applied to our task. 
As described in the main paper, we first experimented with the GRU (textual embedding) of the cross-modal models to describe the scene text and compare it to the captions. Their results are shown in Table~\ref{megatable_v2}, rows (5-8). In contrast to the visual model, where VSRN consistently outperformed VSE++, for scene text the later performs better than the former. Models trained on Flickr30K + TextCaps also perform better than their counterparts trained on Flickr30K only.

We also experimented with training a GRU for a caption-to-scene-text retrieval in Flickr30K. We directly applied the training code of VSE++ to these two modalities (scene text and captions) and simulated the scene text of an image as the intersection between two of its captions. The results of this method, called GRU++, are presented in row (9).

Using GRU trained for cross-modal retrieval (CMR) as scene-text descriptors has its limitations. The scene text is described with a descriptor learned to represent captions, which is not optimal. For scene text, the order of the words is not as relevant as for a caption. However, since the CMR models use a GRU, the scene-text representation is dependent on the order their words are fed to the model.
The Fasttext+FV baseline aims to address these limitations. FastText~\cite{bojanowski2017enriching} uses a larger vocabulary than other Word2Vec based models, and uses word n-grams to embed words. In this manner, FastText is a more robust embedding that learns the syntax as well as the semantics of a given word. On top of FastText, a Fisher kernel~\cite{perronnin2007fisher} is employed to aggregate word embeddings. Additionally, 
an advantage of such an approach is that the scene-text instances are not order dependent and the only training required is at the moment of constructing a Gaussian Mixture Model (GMM) that models the FastText vocabulary distribution. 
The best performing implementation of Fasttext+FV approach is presented in row (11).
On top of it, we show in row (10) a first implementation of this method before lemmatisation and removal of stop words.

Finally, we show results for the two best models (two different flavors of VSE++ GRU) when using OCR prediction from ~\cite{GoogleOCR2020} in rows (5') and (6'). These models are also used in combination with visual-only baselines in rows (19-21), (34-36) and (41-43). We observe a considerable decline in performance between (5) and (5'), (6) and (6'). This can be attributed to errors in OCR prediction. Indeed, COCO-Text is a very challenging dataset for scene-text recognition due to its many small bounding boxes, and CTC inherits these annotations. These results highlights the important of good scene-text recognition for StacMR.
When comparing combinations to their equivalents with ground-truth annotations, the decline in performance is less pronounced.

\textbf{Models trained on Flickr30K} In the main paper, we highlighted how the best performance are obtained from cross-modal retrieval models trained on Flickr30K+TextCaps. We recommend models trained on this combination of datasets for benchmark on CTC. For completeness, we include here re-ranking results for combining models trained on Flickr30K only. Their performance are shown in rows (12-18) using ground-truth scene-text annotations and rows (19-21) using OCR predictions from~\cite{GoogleOCR2020}. 
In comparison to the models trained on Flickr30K+TextCaps, models trained on Flickr30K obtain similar improvement on CTC-1K and more significant gains on CTC-5K.

In addition to these, a few hybrid models (where visual-only models are trained on F30K+TC and scene-text-only models are trained on F30K) are shown in rows (30-36).

\subsection{Performance on TextCaps}
In order to describe why TextCaps is not fit as an evaluation dataset for StacMR, we performed similar experiments to those described in Sections 5.1 of the main paper. The main results are shown in Table~\ref{tab:textcaps}. Here we see how a model trained for cross-modal retrieval with no access to the scene-text information performs better as a scene-text model than a visual model. This highlights the bias of the dataset towards scene text as its main information and the fact that purely visual information comes second.

\subsection{Oracle Late Fusion} \label{sec:oracle}
In addition to providing strong multimodal baselines from separated visual and scene-text models, combination methods are very intuitive to understand. For example, late fusion scores of two models consists of a linear combination of the scores given by two different models. The hyper-parameter $\alpha$ corresponds to the best linear combination factor when averaging for all queries, both images and captions. 

A natural extension to the late fusion combination is to make $\alpha$ a parameter dependent on the values of the the image-to-caption similarity $s_v(q,d)$ and the scene-text-to- caption score $s_t(q,d)$. Based on this extension, we propose an oracle combination method $s_{LF}^{\star}$, called \textit{oracle late fusion}, where the parameter $\alpha$ is query dependent and hand-picked to optimize the ranking for the query. More precisely, this oracle optimizes the median rank of the first retrieved positive item:
\begin{align}
    s_{LF}^{\star}(q,d) &= \alpha^\star (q) s_v(q,d) + (1-\alpha^\star (q))s_t(q,d), \\
    \alpha^\star (q) &= \argmin_{\alpha \in [0,1]}\left( \rank s_{LF}(q,d) \right),
\end{align}
where $\rank$ denotes the rank of the first retrieved positive item. 
Given a visual-only and a scene-text-only model, the oracle late fusion provides us with a theoretical upper-bound to the performance of any combination obtained by linearly combining these models. Moreover, we can analyse the values of $\alpha$ obtained for each query to understand how often does a combination prefers to use the visual model or the scene-text model. Indeed, $\alpha^\star(q)\sim 1$ indicates that, for this query, the visual model is enough and the scene text should be ignored, $\alpha^\star(q)\sim 0$ means that the scene text is enough, and $\alpha^\star(q)$ in between implies a balanced optimal weighting of both modalities.

\begin{figure*}[t!]
  \centering
  \subfloat{\label{ctc_hist1}\includegraphics[width=0.25\textwidth]{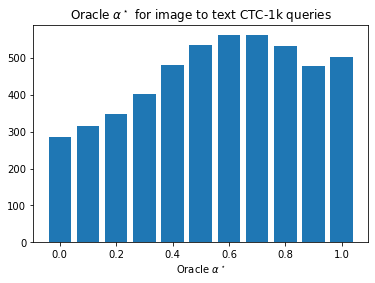}}
  \subfloat{\label{ctc_hist2}\includegraphics[width=0.25\textwidth]{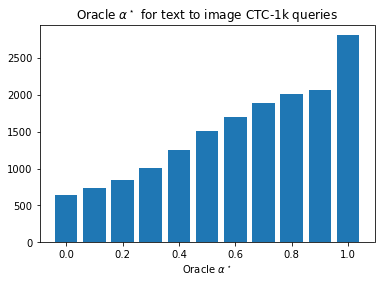}}
  \subfloat{\label{ctc_hist3}\includegraphics[width=0.25\textwidth]{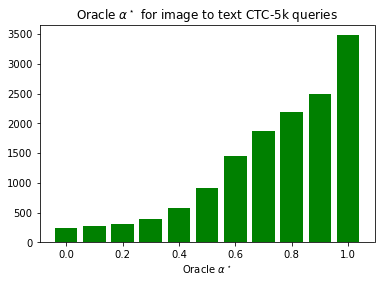}}
  \subfloat{\label{ctc_hist4}\includegraphics[width=0.25\textwidth]{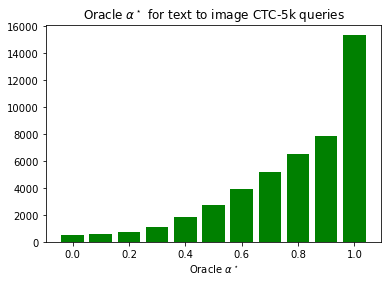}}
  \caption{Histogram of $\alpha$ values for oracle late fusion, row (36) of Table~\ref{megatable_v2}. Blue histograms show oracle $\alpha$ for CTC-1k, green histograms for CTC-5k.}
  \label{oracle_hist}
\end{figure*}

We present the performance for oracle late fusion, evaluated both for CTC and TextCaps, on Table~\ref{megatable_v2} rows (33) and (40), and Table~\ref{tab:textcaps} row (10). We observe a considerable improvement compared to combination methods. While for instance, looking at $R@10$ results, row (39) improved upon row (4) by 4.7$\%$, 2.4$\%$, 1.4$\%$ and -0.3$\%$, row (40) beats row (39) by 10.4$\%$, 10$\%$, 7.8$\%$ and 8$\%$.
More importantly, these theoretical upper-bounds show the unexplored potential of combining visual and scene-text information to improve StacMR results.
We also provide, for the oracle late fusion of row (40), the histogram of optimal values of $\alpha^\star$ in Figure~\ref{oracle_hist}. We observe that $\alpha^\star(q)\sim 1$ more common for text queries than image queries and more common for CTC-5k than CTC-1k. Indeed, text queries and CTC-5k queries have a higher probability to have a zero-word intersection between groundtruth scene text and positive captions, respectively, then image queries and CTC-1k queries, which favors $\alpha^\star=1$.

\section{The STARNet Model} 

\subsection{Implementation Details}
In the baselines of supervised models, SCAN~\cite{lee2018stacked} and VSRN~\cite{li2019visual} use the same hyper parameters as the correspondent work published and it is based on public code available. We introduce modifications to each of those models, in a way that scene-text instances are treated similarly to visual regions. We expanded the number of visual region inputs from the original $36$ to add $15$ scene-text instances that sum in total $51$ combined visual and textual regions. Text instances are sorted according to the confidence value. If text is not present, or the instances are less than $15$, we use a zero-padding scheme.

The proposed supervised model, STARNet was trained for $30$ epochs along with a batch size of $128$ samples per iteration on each experiment. The learning rate employed was $0.0002$ and was decreased by a factor of $10$ every $10$ epochs. 
The visual features have a dimension of $2048$-d. The FastText~\cite{bojanowski2017enriching} textual vectors that serve as input to the model have a dimension of $300$-d, which are linearly projected into a similar feature space of $2048$-d as the visual features. We use $4$ GCN-based reasoning layers on the visual and textual GCN to enrich and reason from the visual and scene-text features. The final semantic space learned contains $2048$-d, which is used to project the final image representation and captions.

In our experiments, when the Flickr30K~\cite{young2014image} dataset is employed, we use the same training, validation and testing split as in~\cite{karpathy2015deep}, which contain $28,000$, $1,000$ and $1,000$ images respectively. When using only the TextCaps~\cite{sidorov2020textcaps} dataset, the original training set is used and the validation set is employed as the evaluation set, since the test set is currently publicly unavailable. At the moment of training the proposed STARNet model, we employ the validation set of TextCaps to achieve the best performing weights.

\subsection{Performance on Flickr30K and TextCaps}

In Table~\ref{tab:supervised_full_results_flickr_ctc} we show the performance of our proposed model with SCAN~\cite{lee2018stacked} and VSRN~\cite{li2019visual}. In order to obtain comparable results, we have obtained features from our implementation to extract visual regions as~\cite{anderson2018bottom}. Publicly available code for SCAN~\cite{lee2018stacked} and VSRN~\cite{li2019visual} was used to train those models. 

Results show that by leveraging scene-text retrieval improvements can be achieved. It is important to note the effect of employing different datasets in the training procedure. As it is expected, training on TextCaps and due to the dataset nature that focuses only on scene text instances, as well as their captions, it does not yield good results when used alone. Even adding samples from the CTC dataset at training time, can yield an improvement when evaluated on the TextCaps validation set.

It is worth noting as well that in standard cross-modal retrieval models, adding TextCaps training data achieve a minor improvement (SCAN) or lower the performance (VSRN) when compared in the Flickr30k dataset. However a slight improvement is achieved when adding the CTC training set. 

However, the proposed model learns to model the interactions between scene-text and visual descriptors to combine them appropriately. STARNet achieves better a performance among both datasets even when scene-text is not widely available in Flickr30k.


\section{Dataset Samples}
 
\begin{figure*}[h]
  \setlength\extrarowheight{5pt} 
  \small
\begin{tabularx}{\linewidth}{@{\extracolsep{\fill}} cX}
  \toprule
  \textbf{Image} & \multicolumn{1}{c}{\textbf{Captions}} \\
\midrule
\multirow{5}{*}{\includegraphics[width=0.2\textwidth, height = 2.7cm]{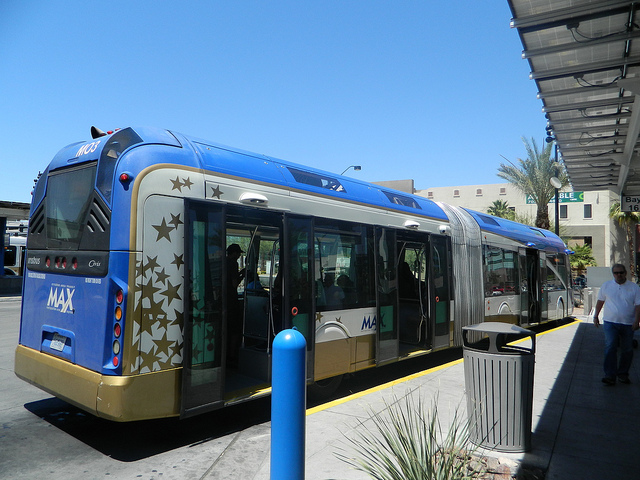}}
& {A blue bus at a bus stop with its doors open.} \\
& {A bus with its doors open is waiting at a bus stop.} \\
& {A bus sits parked on the side of a street.} \\
& {A picture of a bus on the side of the street.} \\
& {The blue and white trolley is waiting on passengers.} \\
\midrule 
\multirow{5}{*}{\includegraphics[width=0.2\textwidth, height = 2.7cm]{images/supp_material_images/full/COCO_train2014_000000327805.jpg}}  
 & \small{A woman, man and two dogs in an inflatable raft on some water.} \\
 & \small{The two ladies are in the row boat.} \\
 & \small{Three people in a raft on the lake.} \\
 & \small{A boat with people on it with a dog in water with a goose in it.} \\
 &\small{Man and woman with two dogs on a power boat on a lake.} \\
\midrule 
\multirow{5}{*}{\includegraphics[width=0.2\textwidth, height = 2.7cm]{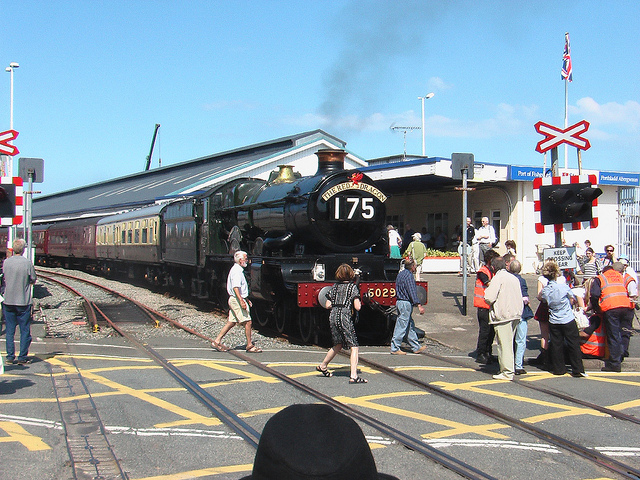}} 
 & \small{A train on the tracks with people standing and walking by it} \\
 & \small{A crowd of people are walking in front of a train} \\
 & \small{A stopped train at a train crossing with people crossing the tracks.} \\
 & \small{A black train parked at a train station as people walk across the train tracks.} \\
 &\small{People at a train station, gathering around a black locomotive.} \\
 
 \midrule 
\multirow{5}{*}{\includegraphics[width=0.2\textwidth, height = 2.7cm]{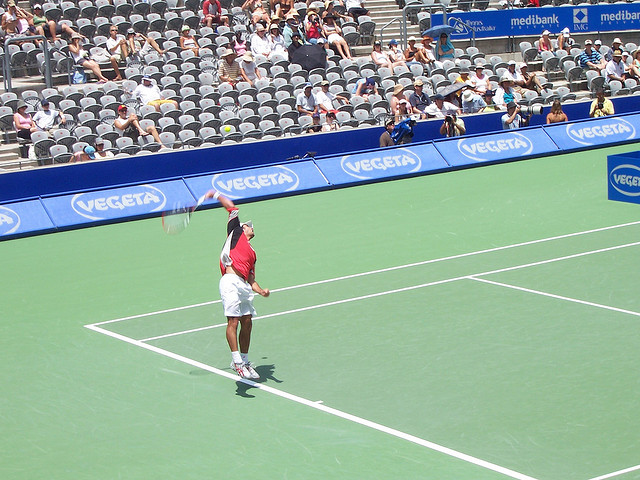}} 
 & \small{A man holding a tennis racquet on a court.} \\
 & \small{A man swinging a tennis racket during a tennis match.} \\
 & \small{A tennis player in mid air action on the court.} \\
 & \small{A tennis player about to serve the ball as a small crowd looks on.} \\
 &\small{A tennis player is in the air making an overhead swing.} \\
 
 \midrule 
\multirow{5}{*}{\includegraphics[width=0.2\textwidth, height = 2.7cm]{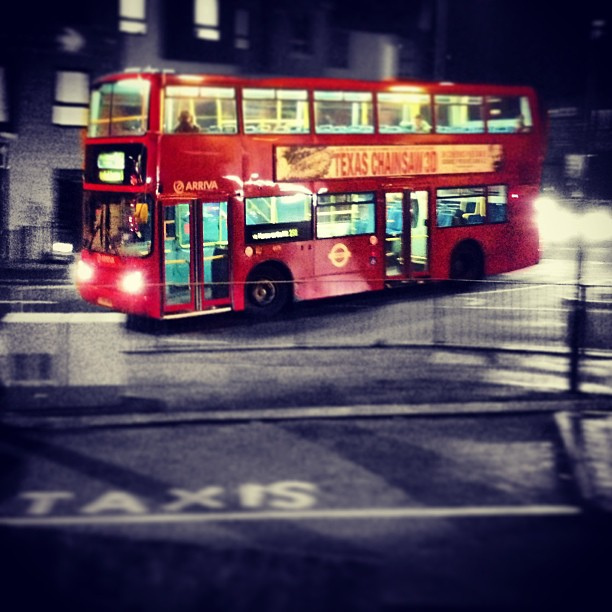}} 
 & \small{A red double decker bus on street next to building.} \\
 & \small{A bus that is driving in the street.} \\
 & \small{A ride double-decker bus stands out against a black and white background.} \\
 & \small{A double decker bus with few passengers turning at a corner.} \\
 &\small{A red double decker bus driving down a city street.} \\
 
\bottomrule
\end{tabularx}
\caption{Image-caption pairs taken from the full proposed CTC dataset, in which appearing scene-text does not have a semantic relation with the annotated captions, \ie there are no scene-text and captions common words.
}
\label{fig:full_samples}
\end{figure*}

\begin{figure*}[t]
  \setlength\extrarowheight{5pt} 
  \small
\begin{tabularx}{\linewidth}{@{\extracolsep{\fill}} cX}
  \toprule
  \textbf{Image} & \multicolumn{1}{c}{\textbf{Captions}} \\
\midrule
\multirow{5}{*}{\includegraphics[width=0.2\textwidth, height = 2.7cm]{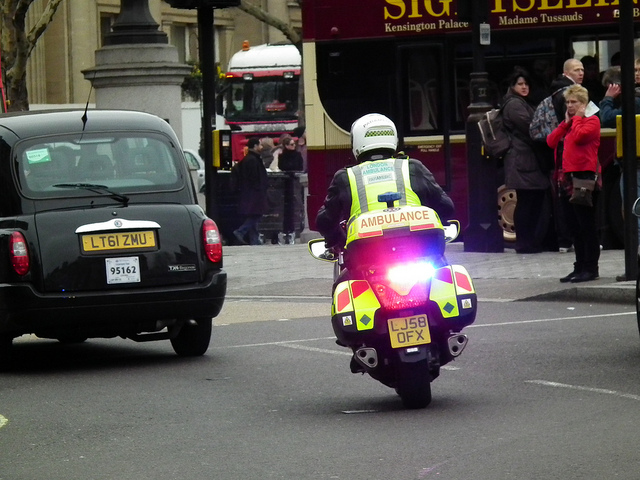}}
& {An emergency response person is on a motorcycle.} \\
& {A medical person riding a motorcycle with \textbf{ambulance} on back.} \\
& {A police officer on a motorcycle pulling over a black car.} \\
& {A police motorcycle gets down to business when someone speeds.} \\
& {A motorcycle with a sign on the back that says \textbf{ambulance}.} \\
\midrule 
\multirow{5}{*}{\includegraphics[width=0.2\textwidth, height = 2.7cm]{images/supp_material_images/explicit/COCO_train2014_000000197120.jpg}}  
 & \small{A \textbf{China Airlines} Airplane sitting on a waiting area of an airport.} \\
 & \small{A big commuter plane sits parked in a air port.} \\
 & \small{A \textbf{China Airlines} airliner is parked at an airport near another jet.} \\
 & \small{Some white red and blue jets at an airport.} \\
 &\small{\textbf{China} airplane airline is parked at a dock.} \\
\midrule 
\multirow{5}{*}{\includegraphics[width=0.2\textwidth, height = 2.7cm]{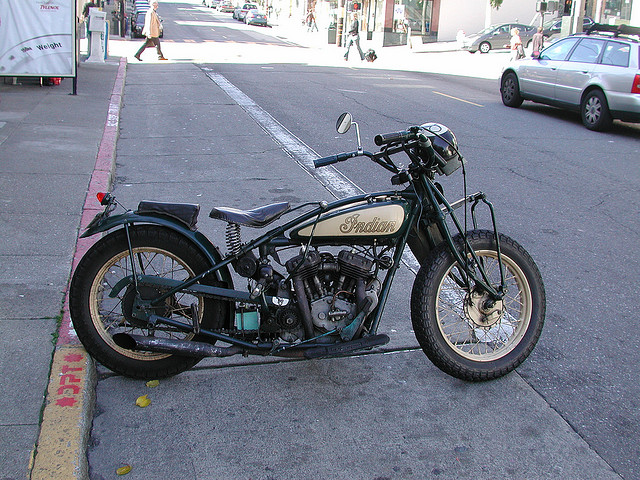}} 
 & \small{A motorcycle parked in a parking lot next to a car.} \\
 & \small{An antique \textbf{Indian} motorcycle is parked next to the sidewalk.} \\
 & \small{Motorcycle parked on the edge of a street.} \\
 & \small{An old \textbf{Indian} motorcycle parked at the curb of a street.} \\
 &\small{A motorcycle parked on a sidewalk next to a street.} \\
 
  \midrule 
\multirow{5}{*}{\includegraphics[width=0.2\textwidth, height = 2.7cm]{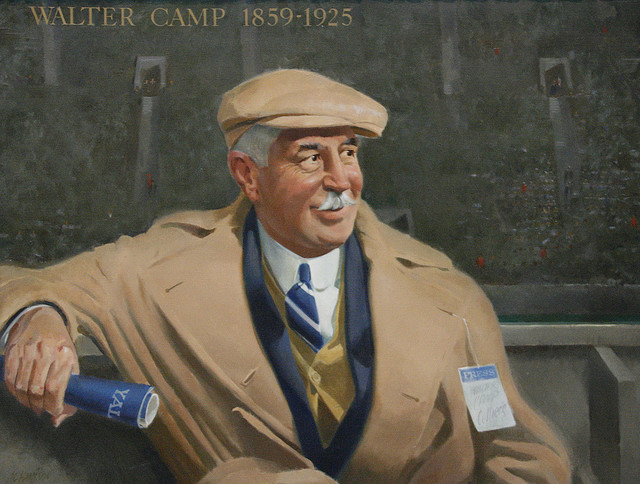}} 
 & \small{Looks like a portrait of a distinguished gentleman.} \\
 & \small{A painting of \textbf{Walter Camp}, siting on bench.} \\
 & \small{A painting of a man in brown jacket and hat sitting at a bench.} \\
 & \small{This a painting of \textbf{Walter Camp} in a trench coat.} \\
 &\small{A painting of an older man on a city bench holding a rolled up magazine.} \\
 
 \midrule 
\multirow{5}{*}{\includegraphics[width=0.2\textwidth, height = 2.7cm]{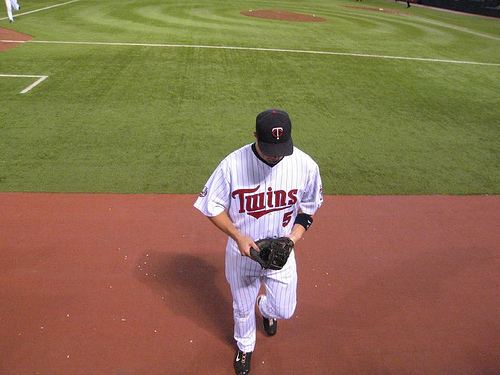}} 
 & \small{A professional baseball player standing on the field while holding a mitt.} \\
 & \small{A baseball player wearing a catchers mitt on top of a field.} \\
 & \small{A \textbf{Twins} baseball player holding his glove walking on the field.} \\
 & \small{The pitcher is resigned to losing the important game.} \\
 &\small{A \textbf{Twins} baseball player walking to the dugout.} \\

\bottomrule
\end{tabularx}
\caption{Image-caption pairs from the proposed CTC explicit dataset, \ie the
  scene-text and captions have at least one word in common (marked in \textbf{bold}).
}
\label{fig:explicit_samples}
\end{figure*}

Figure~\ref{fig:full_samples} showcases a few samples of image-caption pairs that belong to the full CTC dataset. On the other hand, in Figure~\ref{fig:explicit_samples} we depict image-caption pairs that belong to the explicit set of the CTC dataset, the bold words in captions reference to appearing scene text. We can note that scene text provides strong cues to better discriminate each image. Leveraging scene-text can provide with important complementary information for language and vision oriented tasks, such as in the case of cross-modal retrieval.

\section {Qualitative Results}
\begin{figure*}[t]
  \setlength\extrarowheight{5pt} 
  \small
\begin{tabularx}{\linewidth}{@{\extracolsep{\fill}} cX}
  \toprule
  \textbf{Queried Image} & \multicolumn{1}{c}{\textbf{Retrieved Captions}} \\
\midrule
\multirow{5}{*}{\includegraphics[width=0.2\textwidth, height = 2.7cm]{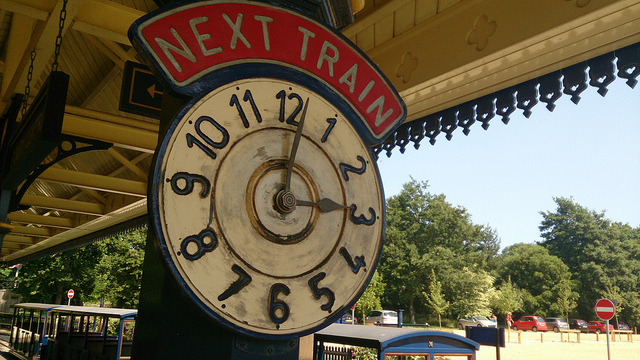}}
& {Clock at a \textbf{train} station showing the time of the next trains arrival. \cmark} \\
& {A clock with the words \textbf{next train} written about it. \cmark} \\
& {A clock on a \textbf{train} platform during day time. \xmark} \\
& {A clock attached to a pole at a \textbf{train} station. $\dagger$ \xmark} \\
& {A clock that is sitting on the side of the pole. \cmark} \\
\midrule 
\multirow{5}{*}{\includegraphics[width=0.2\textwidth, height = 2.7cm]{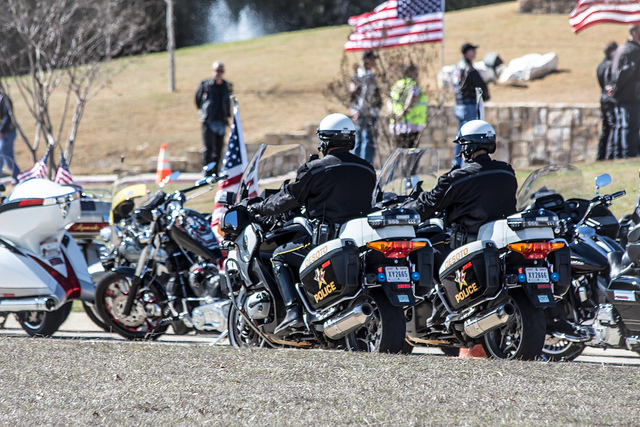}}
& {A large number of \textbf{police} motorcycles are lined up. $\dagger$ \xmark}\\
& {A bunch of \textbf{police} officers on motorcycles waiting for something. \cmark} \\
& {A group of \textbf{police} officers that are riding on motorcycles. $\dagger$ \xmark} \\
& {A \textbf{police} on motorcycles are parked beside a crowd. $\dagger$ \xmark } \\
& {A line of \textbf{police} are riding motorcycles down the street.  \xmark} \\
\midrule 
\multirow{5}{*}{\includegraphics[width=0.2\textwidth, height = 2.7cm]{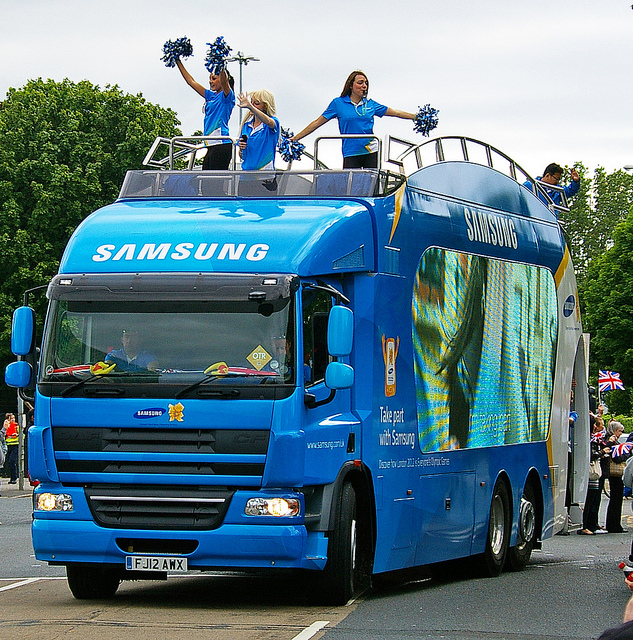}} 
  & \small{People riding on the upper level of a \textbf{samsung} bus in a parade. \cmark} \\
 & \small{A blue tow truck carrying a boat. \xmark} \\
 & \small{A blue truck is pulling a white boat. \xmark} \\
 & \small{A police vehicle on a tow truck that is being taken away. \xmark} \\
 &\small{A group of police standing at the back of a moving truck. \xmark} \\
 
  \midrule 
\multirow{5}{*}{\includegraphics[width=0.2\textwidth, height = 2.7cm]{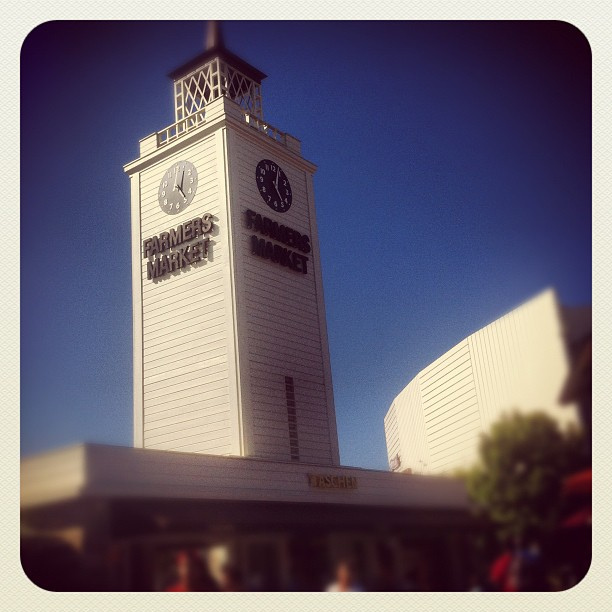}} 
 & \small{A tall lighthouse sign with a clock on the tower of a plaza. \cmark} \\
 & \small{A tall church building with a massive clock on front of it. \xmark} \\
 & \small{A modern clock tower is embellishing a \textbf{market} which sits beneath a clear blue sky. \cmark} \\
 & \small{Tall tower with clock near well lit building at night. \xmark} \\
 &\small{A large tower that has a clock on the very top of it. $\dagger$ \xmark} \\
 
 \midrule 
\multirow{5}{*}{\includegraphics[width=0.2\textwidth, height = 2.7cm]{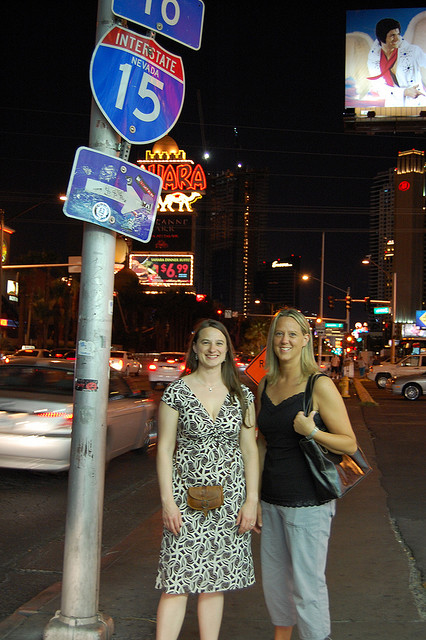}} 
 & \small{Two woman near the interstate \textbf{15} sign in las vegas. \cmark} \\
 & \small{Two women standing on a sidewalk next to a street sign at night while cars drive on the street next to them and behind them. \cmark} \\
 & \small{Two young ladies standing on the sidewalk under a street sign. \cmark} \\
 & \small{Two people standing on a street with a street sign. \cmark} \\
 &\small{Two women on street next to cars and traffic signs. \cmark} \\

\bottomrule
\end{tabularx}
\caption{Qualitative samples obtained when an image is used as a query (Image to Text) in the proposed CTC explicit dataset. Correct results are marked with \cmark. Incorrect results are marked with \xmark. Reasonable mismatches are depicted with $\dagger$ but still marked by a \xmark.
}
\label{fig:qualitative_image2text}
\end{figure*}

In Figure~\ref{fig:qualitative_image2text} we illustrate qualitative results when performing Image to Text cross-modal retrieval. Text contained within an image usually serve as discriminatory signal, such as the word \textit{"samsung"} in the third image and the number \textit{"15"} in the fifth query. Scene text also provides a strong complementary cue to be used along with visual features as the rest of the queried samples suggest.

It is important to note, that even though some samples are not entirely correct, the model still preserves semantics between image and retrieved captions.

\begin{figure*}[h]
    \centering
    \begingroup
    \setlength{\tabcolsep}{2pt} 
    \renewcommand{\arraystretch}{1} 
    \setlength{\fboxsep}{1pt}
    \begin{tabular}{c c c c c}
    \multicolumn{5}{c}{\fontfamily{lmss}\selectfont Query 1: \emph{A \textbf{marc} passenger drains rides along railroad tracks.}} \\
    \colorbox{red}{\includegraphics[width=0.17\textwidth, height=0.17\textwidth]{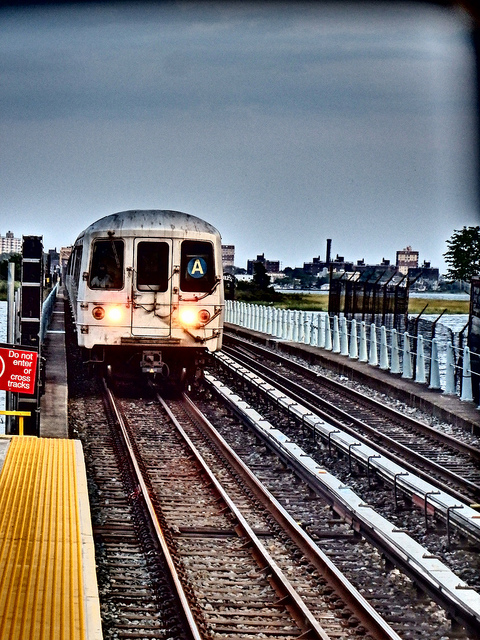}} &
    \colorbox{red}{\includegraphics[width=0.17\textwidth, height=0.17\textwidth]{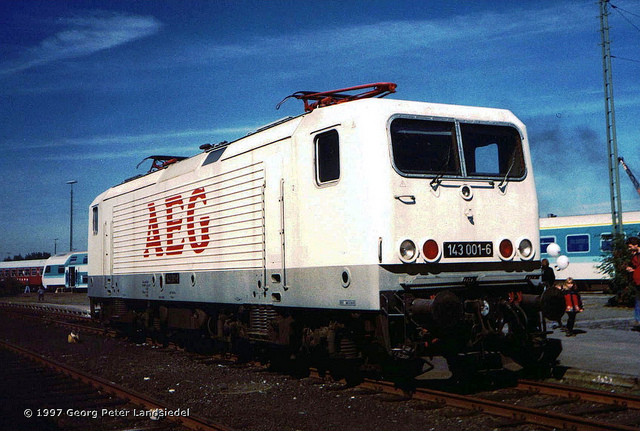}} &
    \colorbox{green}{\includegraphics[width=0.17\textwidth, height=0.17\textwidth]{images/supp_material_images/qualitative_text2image/q1_3.jpg}} &
    \colorbox{red}{\includegraphics[width=0.17\textwidth, height=0.17\textwidth]{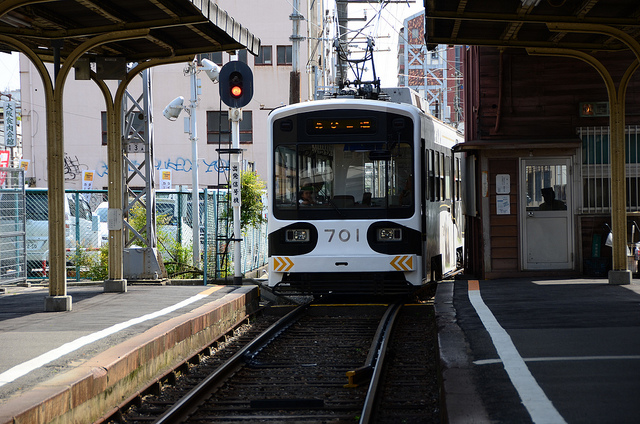}} &
    \colorbox{red}{\includegraphics[width=0.17\textwidth, height=0.17\textwidth]{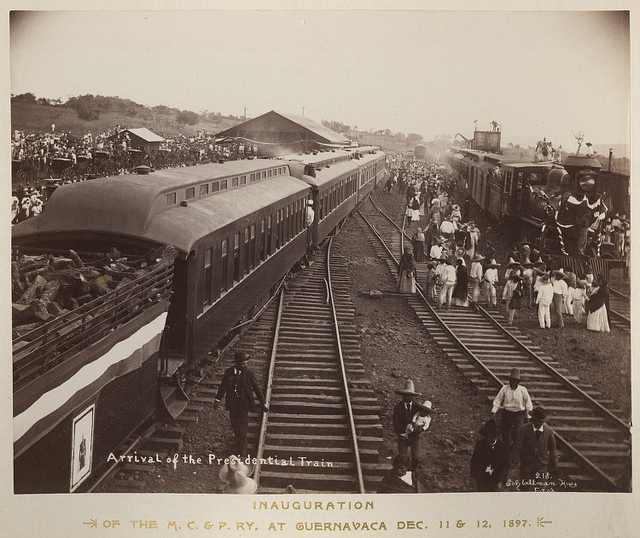}} \\
    \multicolumn{5}{c}{}\\ 
    \multicolumn{5}{c}{\fontfamily{lmss}\selectfont Query 2: \emph{Sign explaining how to \textbf{park} on a hill is posted on the street.}} \\
    \colorbox{green}{\includegraphics[width=0.17\textwidth, height=0.17\textwidth]{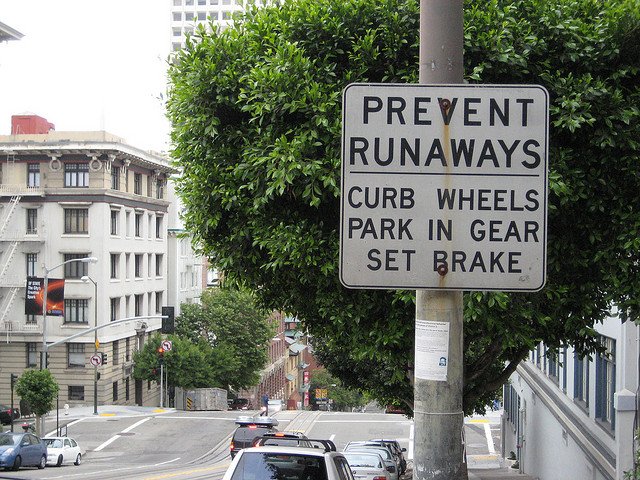}} &
    \colorbox{red}{\includegraphics[width=0.17\textwidth, height=0.17\textwidth]{images/supp_material_images/qualitative_text2image/q2_2.jpg}} &
    \colorbox{red}{\includegraphics[width=0.17\textwidth, height=0.17\textwidth]{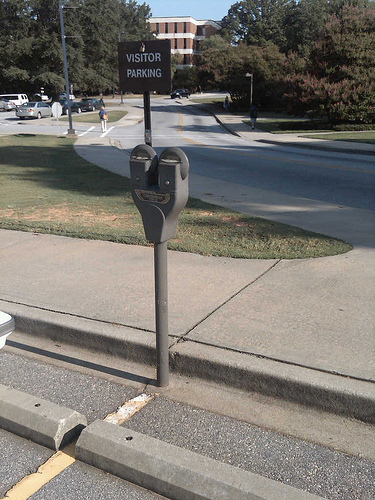}} &
    \colorbox{red}{\includegraphics[width=0.17\textwidth, height=0.17\textwidth]{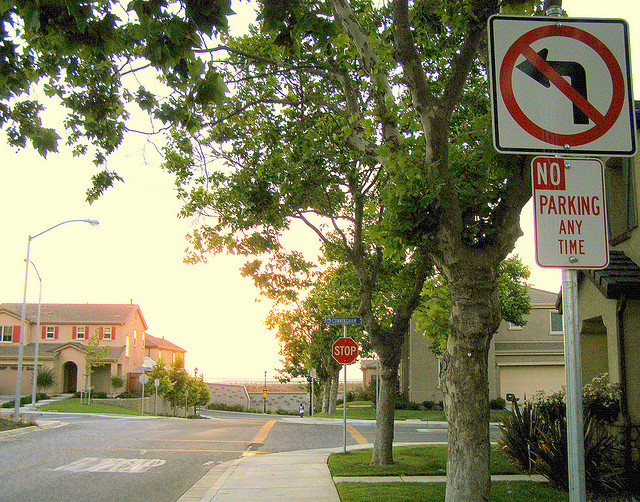}} &
    \colorbox{red}{\includegraphics[width=0.17\textwidth, height=0.17\textwidth]{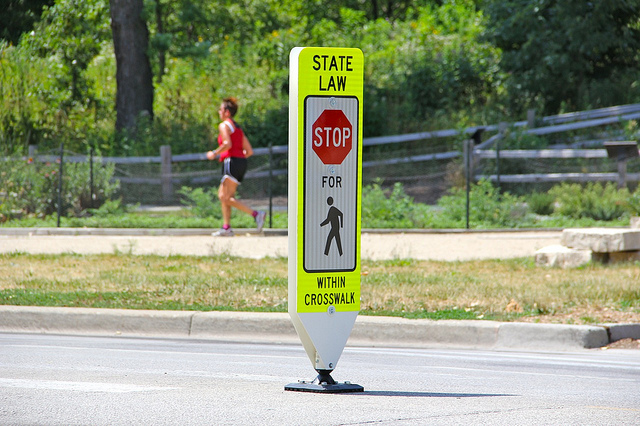}} \\
    \multicolumn{5}{c}{}\\ 
    \multicolumn{5}{c}{\fontfamily{lmss}\selectfont Query 3: \emph{Commuter \textbf{shuttle} bus on roadway in large city.}} \\
    \colorbox{red}{\includegraphics[width=0.17\textwidth, height=0.17\textwidth]{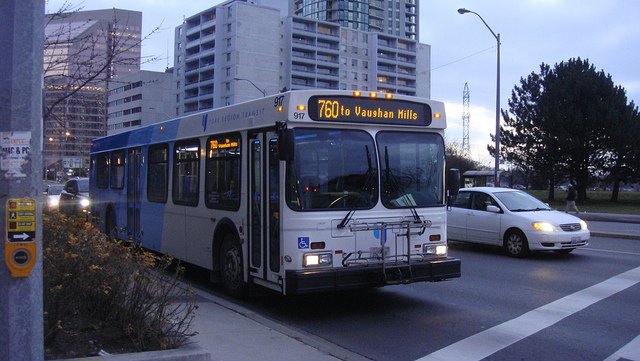}} &
    \colorbox{green}{\includegraphics[width=0.17\textwidth, height=0.17\textwidth]{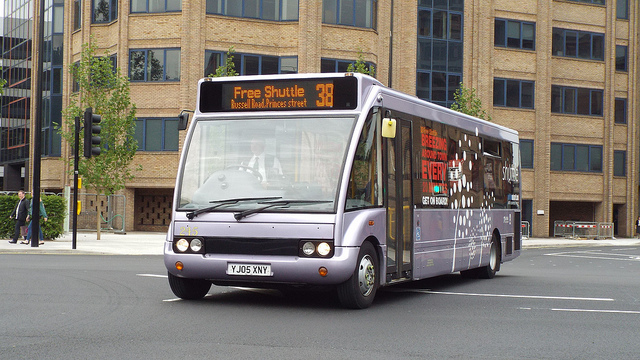}} &
    \colorbox{red}{\includegraphics[width=0.17\textwidth, height=0.17\textwidth]{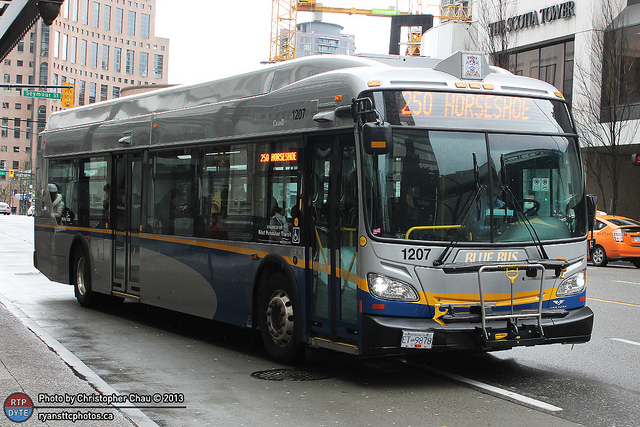}} &
    \colorbox{red}{\includegraphics[width=0.17\textwidth, height=0.17\textwidth]{images/supp_material_images/qualitative_text2image/q3_4.jpg}} &
    \colorbox{red}{\includegraphics[width=0.17\textwidth, height=0.17\textwidth]{images/supp_material_images/qualitative_text2image/q3_5.jpg}} \\
    \multicolumn{5}{c}{}\\ 
    \multicolumn{5}{c}{\fontfamily{lmss}\selectfont Query 4: \emph{A \textbf{china airlines} airliner is parked at an airport near another jet.}} \\    
    \colorbox{green}{\includegraphics[width=0.17\textwidth, height=0.17\textwidth]{images/supp_material_images/qualitative_text2image/q4_1.jpg}} &
    \colorbox{red}{\includegraphics[width=0.17\textwidth, height=0.17\textwidth]{images/supp_material_images/qualitative_text2image/q4_2.jpg}} &
    \colorbox{red}{\includegraphics[width=0.17\textwidth, height=0.17\textwidth]{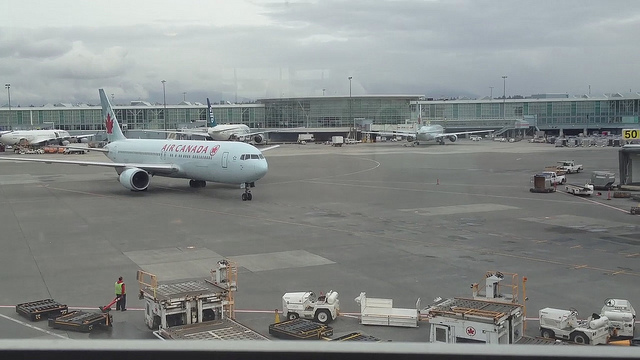}} &
    \colorbox{red}{\includegraphics[width=0.17\textwidth, height=0.17\textwidth]{images/supp_material_images/qualitative_text2image/q4_4.jpg}} &
    \colorbox{red}{\includegraphics[width=0.17\textwidth, height=0.17\textwidth]{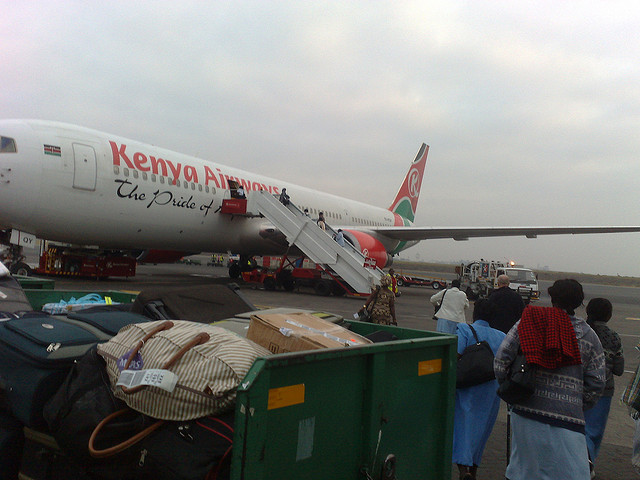}} \\
    \end{tabular}
    \endgroup
    \caption{Qualitative samples when a caption is used as a query (Text to Image) in the proposed CTC explicit dataset. Correct results are marked in a green box. Incorrect results are marked in a red box. Words in bold in queried captions depict the scene-text that helps to discriminate retrieved images, which otherwise are ambiguous. Query 1 contains an annotator typo "drains".}
    \label{fig:qualitative_text2image}
\end{figure*}

We illustrate in Figure~\ref{fig:qualitative_text2image} the results obtained when performing Text to Image cross-modal retrieval. In the queries performed, scene-text work as fine-grained and discriminative information to retrieve correctly an image. Similarly to the previous scenario, wrongly retrieved samples still preserve semantics. 

By exploring the qualitative results obtained, added to the quantitative tables in previous sections, we can reinforce the notion that modelling scene-text along with visual features does improve retrieval granularity thus yielding higher performing cross-modal retrieval pipelines. 

\end{document}